\documentclass{article}

\usepackage{arxiv}

\usepackage[utf8]{inputenc} 
\usepackage[T1]{fontenc}    
\usepackage{hyperref}       
\usepackage{url}            
\usepackage{booktabs}       
\usepackage{amsfonts}       
\usepackage{nicefrac}       
\usepackage{microtype}      
\usepackage{lipsum}		    
\usepackage{graphicx}
\usepackage{natbib}
\usepackage{doi}
\usepackage{caption}
\usepackage{subcaption}
\usepackage{orcidlink}

\title{Unified Deep Learning Model for Global Prediction of Aboveground Biomass, Canopy Height and Cover from High-Resolution, Multi-Sensor Satellite Imagery}

\author{Manuel Weber\orcidlink{0000-0002-9926-7914}\thanks{Corresponding author}\hspace{1cm}Carly Beneke\hspace{1cm}Clyde Wheeler \vspace{3mm} \\
	Descartes Labs Inc. \\
        Santa Fe, NM, USA
}

\hypersetup{
pdftitle={Unified Deep Learning Model for Global Prediction of Aboveground Biomass, Canopy Height and Cover from High-Resolution, Multi-Sensor Satellite Imagery},
pdfsubject={cs.AI, cs.CV, cs.LG},
pdfauthor={Manuel Weber},
pdfkeywords={Biomass, Canopy height, Canopy cover, Deforestation, Carbon accounting, Remote sensing, LiDAR, GEDI, Deep learning, Sustainability},
}

\begin{document}
\maketitle

\begin{abstract}
Regular measurement of carbon stock in the world's forests is critical for carbon accounting and reporting under national and international climate initiatives, and for scientific research, but has been largely limited in scalability and temporal resolution due to a lack of ground based assessments. Increasing efforts have been made to address these challenges by incorporating remotely sensed data. We present a new methodology which uses multi-sensor, multi-spectral imagery at a resolution of 10 meters and a deep learning based model which unifies the prediction of above ground biomass density (AGBD), canopy height (CH), canopy cover (CC) as well as uncertainty estimations for all three quantities. The model is trained on millions of globally sampled GEDI-L2/L4 measurements. We validate the capability of our model by deploying it over the entire globe for the year 2023 as well as annually from 2016 to 2023 over selected areas. The model achieves a mean absolute error for AGBD (CH, CC) of 26.1~Mg/ha (3.7~m, 9.9~\%) and a root mean squared error of 50.6~Mg/ha (5.4~m, 15.8~\%) on a globally sampled test dataset, demonstrating a significant improvement over previously published results. We also report the model performance against independently collected ground measurements published in the literature, which show a high degree of correlation across varying conditions. We further show that our pre-trained model facilitates seamless transferability to other GEDI variables due to its multi-head architecture.
\end{abstract}


\section{Introduction}
It is estimated that the worlds forests store up to 500~petagram (Pg) of carbon in their above ground biomass (\cite{fao2020}) and act as a major carbon sink which is vital for maintaining stable ecosystems. According to the IPCC Sixth Assessment Report (\cite{IPCC2023}), deforestation and forest degradation contribute roughly 8-10~\% of global greenhouse gas emissions, accelerating climate change. Regular assessment of global above ground carbon stock is necessary to understand and mitigate climate impacts as well as for accounting for corporate emissions disclosures, regulatory compliance, and is part of international climate initiatives, including the UN Paris Agreement (\cite{UNFCCC_ParisAgreement_2015}). This poses an immense challenge since the only validated and accurate method for both measuring the tree profiles (including canopy height) and calibration of ecosystem specific allometric equations (allowing the conversion of tree profile into biomass) requires labor intense field measurements. More efficient and scalable methods involve air- or space-borne LiDAR instruments which are capable of scanning the tree profiles across large areas (\cite{BALESTRA2024}). However, there is a trade-off between the density and resolution of such LiDAR maps and the spatial extent.
\begin{figure}[h]
	\centering
	\includegraphics[width=\textwidth]{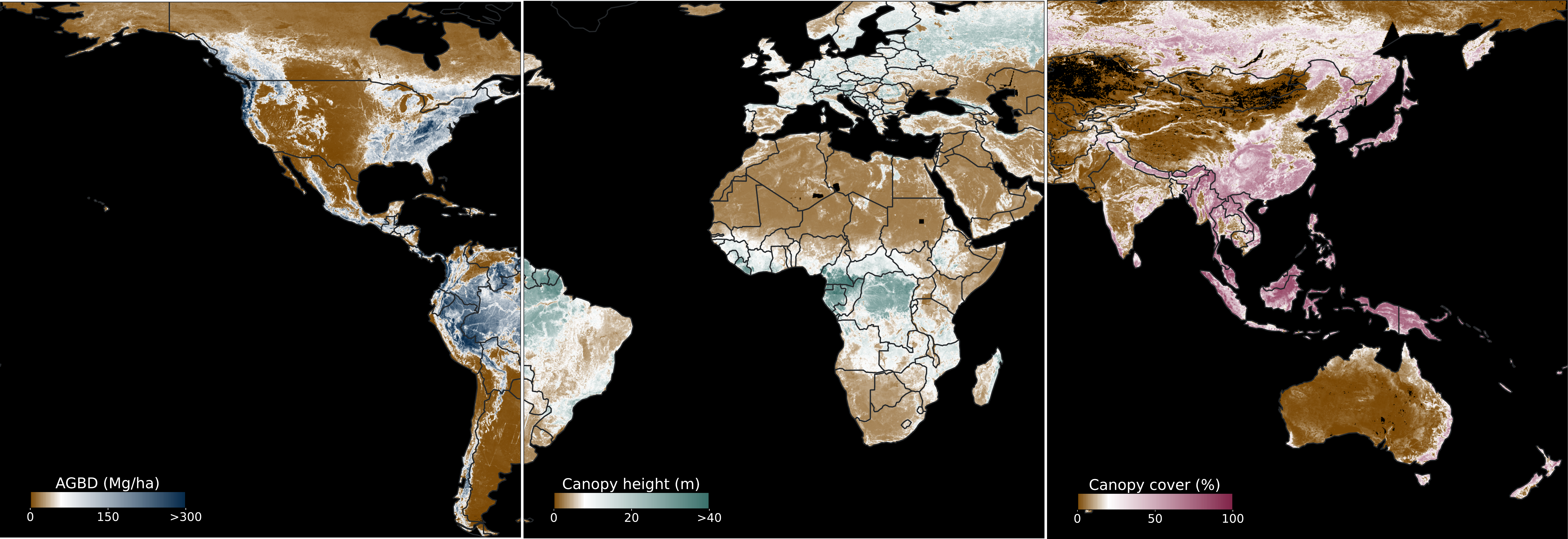}
	\caption{Global maps of aboveground biomass density, canopy height and canopy cover at 10~m resolution, simultaneously estimated by our unified model.}
	\label{fig:global_map_all}
\end{figure}
For example, airborne LiDAR surveys can generate high resolution maps, due to the dense flight paths, but are limited in the area that can be covered. Space-borne instruments such as the Global Ecosystem Dynamics Investigation (GEDI) project (\cite{DUBAYAH2020100002}) can cover the entire globe but their measurements are very sparse. In order to bridge the gap between dense estimation and global scalability, remotely sensed data, together with machine learning techniques, have become a main focus. Early approaches used simple ML methods, such as random forest algorithm, resulting in pixel-wise estimation of canopy height on local scales (\cite{rs15081969}) as well as low resolution global canopy height maps (CHM) (\cite{POTAPOV2021112165}). Recent methods have pushed the boundaries and generated higher resolution maps, both of AGBD (\cite{Bereczky2024}) and CH (\cite{pauls2024estimatingcanopyheightscale}), by combining high resolution imagery with various sources of LiDAR and ground measurements as well as state of the art computer vision models. Despite these advancements, there remain multiple challenges due to the availability of frequent, cloud free observations of high resolution satellites. In many cases, the imagery used spans a long and inconsistent time range which complicates the ability to track changes in carbon cycle. Moreover, many solutions achieve high accuracy for estimating AGBD or CH on regional scales but lack the ability to expand the model to the entire globe. The benefit of a single global model offers the learning from a diverse set of training samples, across multiple ecosystems, resulting in a base model with good accuracy for broad scale monitoring. Further fine-tuning of the base model locally can achieve greater accuracy with far less effort than training multiple local models from scratch. Especially when AGBD is determined using local allometric equations, employing the base model accelerates the workflow and further reduces error metrics.
\\ \\
The detection of deforestation events require the classification of land cover into forest for which different definitions based on CH and CC exist (\cite{EC_Deforestation_Definitions}). Having estimates of both variables at global scale, allows the application of different forest definitions dynamically and detect deforestation events by differentiation of binary maps at different times. In this work, we present a new approach which addresses many of the shortcomings mentioned above by introducing a deep learning based model which unifies the prediction of AGBD, CH and CC as well as their uncertainties into one single model, trained on a global dataset of fused Sentinel-1, Sentinel-2, digital elevation model (DEM) and geographic location. We leverage the data from GEDI to generate sparse ground truth maps. In total, the dataset consists of $\sim$1~M data samples composed of image tiles of size 256~pixels$\times$256~pixels at 10 meters ground distance. In addition, $\sim$60~k data samples were generated as test dataset. Each image tile is a cloud free composite amounting to $\sim$13.8~M total scenes used across all sensors. Each image tile is required to contain $>$20 GEDI footprints resulting in a total of $\sim$67~M ($\sim$3.8~M) ground measurements in the training (test) dataset. Our model consists of a convolutional neural network (CNN) with a pre-defined, trainable encoder network and a custom decoder network to form a feature pyramid network (FPN). Due to the spatially sparse ground truth distribution, we introduce a novel technique for training our model which results in better performance compared to traditional methods. We show the results of the model deployment over the recent (2023) year at global scale as well as historic imagery back to the year 2016 in intervals of 1 year at a local scale. We first outline previous works (section \ref{sec:previous_work}) before describing the dataset created for this work (section \ref{sec:data}) and our methodology including a detailed description of our modeling approach and training procedure (section \ref{sec:methodology}). We then present the results of assessing the model accuracy against a globally sampled test dataset as well as independent ground measurements (section \ref{sec:results}), illustrate applications and use cases, including the result of the global model deployment (section \ref{sec:applications}), and conclude with a discussion of our results (section \ref{sec:discussion}). The main contribution of this work can be summarized as:
\begin{itemize}
    \item First deep learning based model which unifies the prediction of above ground biomass density, canopy height and canopy cover as well as their respective uncertainties
    \item Advancements for predicting global scale maps at high resolution at regular time intervals without missing data due to multi-sensor fusion
    \item Novel training procedure for sparse ground truth labels
    \item Extensive evaluation against third party datasets as well as demonstration of model fine-tuning for local conditions
\end{itemize}

\section{Previous work}
\label{sec:previous_work}
Over the past two decades increasing focus and effort has been directed to environmental monitoring based on space-borne earth observation missions. Such missions go as far back as the 1970s with the Landsat (\cite{WULDER20122}) constellation which offers an immense archive of medium resolution satellite imagery. In recent years, specialized missions have paved the way for more accurate insights into the global dynamics of environments with higher revisit rates and resolution, including multi-spectral passive sensors (\cite{DRUSCH201225}, \cite{DONLON201237}, \cite{VEEFKIND201270}, \cite{IRONS201211}), active sensors such as synthetic aperture radar (SAR) (\cite{TORRES20129}, \cite{MARKUS2017260}) light detection and ranging (LiDAR) (\cite{DUBAYAH2020100002}) and missions dedicated to understanding the carbon cycle (\cite{LETOAN20112850}, \cite{9791979}). The increasing volume of data collected by all these missions has motivated the development of modern and novel algorithms based on machine- and deep learning (\cite{PEHADING2024}). Previous work has focused on the model development for estimating either aboveground biomass density, canopy height or canopy cover. We could not find any references which combine the prediction of multiple variables into a single model. In most previous approaches the limitation in spatial resolution arises from the choice of input data source and ranges from 250~m-1~km (e.g. MODIS) to 30~m (e.g. Landsat), 10~m (e.g. Sentinel-1/2) and $\sim$1~m (e.g. MAXAR).
\\ \\
\textit{Biomass.} Aboveground biomass maps are available today, covering up to decades of history, but are often produced at local scale based on ground measurements and forest inventories (\cite{rs11232744}). Scaling these maps to larger regions requires the collection of plot data which cover various ecosystems. Capturing ground truth data across the broad range of ecosystems and land cover that would be required to scale this methodology would be prohibitively expensive. Early efforts which incorporate simple machine learning techniques have focused on pan-tropical regions and use medium to low resolution satellite imagery (\cite{SAATCHI2011}, \cite{BACCINI2012}, \cite{BACCINI2016}). At global scale, a number of aboveground biomass maps have been generated at low resolution ($\sim$1~km) (\cite{2016AGUFM.B51F0470S}, \cite{9103217}) including the gridded version of the GEDI level-4 product (\cite{DUBAYAH2022}). Only recently, with the incorporation of modern deep learning techniques, higher resolution maps at scale have emerged (\cite{sialelli2024agbdglobalscalebiomassdataset}, \cite{Bereczky2024}) which combine the use of satellite imagery with global scale LiDAR surveys.
\\ \\
\textit{Canopy height.} In contrast to aboveground biomass, canopy height estimations from satellite imagery is less reliant on regional calibrations as ground measurements can be gathered directly from forest inventories or LiDAR measurements which provide insights into the canopy structure. Early approaches utilize simple pixel-to-pixel machine learning algorithms such as random forest at medium resolution (\cite{LI2020102163}, \cite{POTAPOV2021112165}) while more recent methodologies were developed based on deep learning models and higher resolution, single-sensor imagery (\cite{LANG2023}) as well as combining multiple sensors as input data (\cite{pauls2024estimatingcanopyheightscale}). The advancements in deep learning based computer vision models, which exhibit great skills at depth estimation (\cite{oquab2024dinov2learningrobustvisual}), have allowed the development of very high resolution canopy height maps (\cite{TOLAN2024113888}) and models which characterize single trees (\cite{LI2023}, \cite{MUGABOWINDEKWE2023}, \cite{cambrin2024depthcanopyleveragingdepth}). The main drawback of very high resolution maps at global scale is the immense computational effort and cost for a single deployment. In order to cover the entire globe, high resolution imagery is often gathered within a large time window (multiple years) which creates an inconsistency in the temporal resolution and complicates the change monitoring. In order to create consistent and high quality global maps, it is therefore advisable to revert to lower resolution imagery ($\sim$10~m) with higher frequency of observations and to merge multiple sources which may compliment each other. Large scale LiDAR surveys provide a more direct way of generating canopy height maps since they do not rely on models which estimate canopy height based on imagery which has limitations in information content. However, high-resolution aerial surveys (\cite{NEON2021}) are limited in scalability while global scale surveys have low resolution (\cite{DUBAYAH2021}).
\\ \\
\textit{Canopy cover.}  Estimating canopy cover from satellite imagery is the least complex of the three tasks as it does not rely on detailed three dimensional canopy structure or local calibrations. However, it may still pose many challenges due to the quality and resolution of the input imagery. The earliest global canopy cover maps based on remote sensing are derived from Landsat imagery at medium resolution (\cite{HANSEN2013}). Different satellite sources at varying resolutions have been used in generating regional maps (\cite{land10040433}, \cite{s23073394}) while historic imagery, despite the low resolution, allows for high frequency updates reaching back many decades (\cite{LIU2024}).
\\ \\
To the best of our knowledge, our work is the first to combine the estimation of aboveground biomass density, canopy height and canopy cover into a single unified model. With respect to single variable estimation, our approach is most similar to (\cite{Bereczky2024}) for aboveground biomass density and (\cite{pauls2024estimatingcanopyheightscale}) for canopy height which we compare our model evaluation results to (see section \ref{sec:results}).

\section{Data}
\label{sec:data}
In this work, we utilize multi-spectral, multi-source satellite imagery, digital elevation model as well as geographic coordinates as input to the model. The model is trained in a weakly supervised manner (see section \ref{sec:methodology}) on point data from the Global Ecosystem Dynamics Investigation (GEDI) instrument. In this section, we describe the processing steps for generating a global dataset of input and target samples. We will also briefly explain the relevant concepts of the GEDI mission as well as the different data processing levels as it will be an important aspect of understanding the inherent model uncertainties and its limitations.

\subsection{Ground truth data}
\label{sec:ground_truth_data}
The GEDI instrument is a space-borne LiDAR experiment mounted on the International Space Station (ISS) and has been operational since 2019. It comprises 3 Nd:YAG lasers, optics and receiver technology allowing to measure the elevation profile along the orbital track of the ISS. Within the lifetime of the experiment, it is expected to collect 10 billion waveforms at a footprint resolution of 25~m. The setup of 3 lasers, one of which is split into two beams, as well as dithering every second shot leads to a pattern of point measurements with 8 tracks per pass where the tracks are separated by 600~m and each point by 60~m along the flight path. Each GEDI measurement consists of the waveform resulting from the returned signal of the laser pulse sent out at the given location. The collection of all these waveforms is referred to as level-1 data. Each waveform is further processed to extract metrics which characterize the vertical profile of the trees within a given beam footprint.
\\ \\
The signal with the longest time of flight corresponds to the ground return and is used as the reference for the relative height (RH) metrics. The RH[{X}] metrics correspond to the relative height at which [{X}] percent of the total accumulated energy is returned. These metrics characterize the vertical profile of the GEDI footprint where RH100 corresponds to the largest trees in the footprint. These metrics, together with other parameters related to the measurement conditions, are stored in a dataset referred to as level-2a. In additional steps these metrics are used to calculate the percent canopy coverage (level-2b) as well as a gridded version (level-3). Further processing involving regional calibration of allometric equations, leveraging level-2 data, results in estimations of aboveground biomass density (AGBD) as well as uncertainties referred to level-4a. The estimations are based on models which were fit to on-the-ground biomass measurements in a number of field plots located in various regions around the globe. Since most of these measurements do not intersect with a GEDI footprint, airborne LiDAR was used to measure the return signal which was then translated into a simulated GEDI waveform. In this work we utilize both the level-2a/b and level-4a data as ground truth. The on-the-ground measurements for biomass were mostly done without any tree clearing but by measuring canopy height and diameter and using allometric equations, specific to the tree type and world region, to determine biomass. The simulated waveforms undergo the same processing to extract RH metrics which provide the predictors for linear models to predict AGBD. Various models were developed for all combinations of plant functional type (PFT) and world region defined as prediction stratum. For details about the selected predictors for each model and their performances see section 2 and 3 in (\citet{DUNCANSON2022112845}).
\\ \\
The models are linear functions of the predictors with a general form of
\begin{equation}
    \label{eq:gedi_linear_model}
    f_{j}=X_{j}b_{j}
\end{equation}
where $X_{j}$ is a $n \times m$ matrix of $n$ measurements with $m$ predictors and $b_{j}$ a $m \times 1$ vector of parameters for prediction stratum $j$. The best parameters are determined by linear regression where the predicted variable may be in transformed units using a function $h$ which is either unity, square root or log. For new measurements $X'$ the model of the corresponding prediction stratum is chosen and the predictions are given by
\begin{equation}
    \label{eq:gedi_agbd}
    AGBD=h^{-1}\left(X^{\prime}b+\epsilon\right)
\end{equation}
where $\epsilon$ is a bias term determined by the fit residuals in the respective prediction stratum. For each prediction of footprint $k$ a standard error is calculated which is defined as
\begin{equation}
    \label{eq:gedi_se}
    SE_{k}=\sqrt{MSE_{j}+X_{k}\mbox{cov}(b)X_{k}^{T}}
\end{equation}
as well as a confidence interval given by
\begin{equation}
    \label{eq:gedi_agbd_se}
    AGBD\pm t\left(1-\frac{\alpha}{2},n-2\right)\cdot SE_{k}
\end{equation}
where $t$ is the value of the t-distribution with n-2 degrees of freedom at a confidence level of $\alpha$. In general, we observe that the uncertainty on AGBD from the GEDI ground truth is increasing with the value of AGBD where the main contribution comes from the residuals in equation \ref{eq:gedi_se}. This is an important fact to consider when training the model. During model optimization, it is generally assumed that the target values are the absolute truth while in this case we know that the target values are inherently uncertain. This means that a given input $X$ can be mapped to two different values $Y_{1}$ and $Y_{2}$ which can not be described by a continuous function. Since only continuous activation functions are used in our CNN, the function it represents will also be continuous which may lead to predictions with larger uncertainties or under-prediction in regions where the ground truth data is uncertain. We will discuss this further in section \ref{sec:results} and appendix \ref{apx:gedi_uncertainty}. The largest ground truth uncertainty is introduced when converting canopy height, or generally RH metrics, to AGBD which is more accentuated by the lack of ground plot measurements used for calibration. Our model provides an important benefit by not just predicting AGBD in an end-to-end setup, but at the same time also predicts canopy height which can be used a posteriori for re-calculating AGBD should more accurate plot data become available.
\\ \\
The multi-head architecture of our model (see section \ref{sec:model_development}) allows for simultaneous prediction of multiple GEDI variables. For the training of the base model, we choose AGBD from level-4a, CH from level-2a and CC from level-2b as the prediction variables. In section \ref{sec:model_finetuning} we demonstrate how the base model can be easily fine-tuned on different variables in the GEDI dataset. The level-2a dataset provides relative height metrics at discrete energy return quantiles from 0 to 100 in steps of 1, denoted RH00, RH01, ..., RH99, RH100. It is common to choose one of RH95, RH98 or RH100 to define CH. Our base model uses RH98 which prevents over prediction in cases where there is a single or a few very large trees among smaller trees. We save all RH metrics in our datasets such that they can be dynamically selected during training. This provides the flexibility to fine-tune the model on a different RH metric, or multiple RH metrics, depending on the requirement of local allometric equations.

\subsection{Input data}
\label{sec:input_data}
In order to leverage the respective benefits of different data sources, we fuse optical bands (red, green, blue) from the Sentinel-2 (\cite{sentinel2_2021}) satellite with thermal bands (nir, swir1, swir2) from the same satellite (processed with the SEN2COR algorithm (\cite{10.1117/12.2278218}) to provide surface reflectance) and synthetic aperture radar (SAR) signal (VV and VH backscatter) from Sentinel-1 (\cite{9884753}). In addition, we use altitude, aspect and slope from the Shuttle Radar Topography Mission (SRTM) (\cite{nasa_srtm_2013}) to further enrich the predictive capability of our model. The predictors from the digital elevation model (DEM) carry important information about the local topography which affects the distribution of plant functional types  and their growth pattern (\cite{WANG2022152116}). We further provide the global coordinates of each data sample by encoding longitude and latitude in the interval [-1, 1]. The optical bands of Sentinel-2 as well as the Sentinel-1 bands have a native resolution of 10~m while the thermal bands of Sentinel-2 have a resolution of 20~m and the DEM of 30~m. In order for the CNN to process the input layers at different resolutions, we resample all layers to 10~m resolution using the bilinear interpolation method and stack them to form a 13 channel input tensor.
\\ \\
To generate a global training and test dataset, we uniformly sample locations within the latitude (longitude) range of [-51.6, 51.6] degrees ([-180, 180] degrees) which intersect with landmass. We use the Descartes Labs (DL) proprietary tiling system to create image tiles of size 512~pixels$\times$512~pixels at 10~m/pixel resolution with the sampled coordinate being at the center of the tile. Each tile is required to contain >20 GEDI footprints. During training, we dynamically split each tile into 4 non-overlapping sub-tiles of size 256~pixels$\times$256~pixels, which increases the total number of samples in the dataset four-fold. We introduce a naming convention for the tiles based on their location being either in the southern hemisphere (lat $<$ -23.5$^\circ$), the tropics (-23.5$^\circ$ $\leq$ lat $\geq$ 23.5$^\circ$) or the northern hemisphere (lat $>$ 23.5$^\circ$).

\subsubsection{Cloud mask and image composite}
\label{sec:cloud_mask}
In order to reduce obstruction from clouds and cloud shadows in Sentinel-2 imagery, we generate cloud free composites from a stack of images which are masked with the binary output of our proprietary cloud and cloud shadow detection model. The image stack is built by collecting all Sentinel-2 scenes which intersect with the given tile within a specified time range. Scenes taken on the same day are mosaiced. We use the median operation to generate the composite from the masked image stack. The composite time range is chosen to minimize the variability in spectral response from vegetation while maximizing the resulting coverage. We therefore chose the time ranges to be the respective summer months for the two hemispheres (June - August for the northern hemisphere and December - February for the southern hemisphere) except for the tropical region where the composite was done over a 6 months period. In certain tropical regions, cloud artifacts are visible despite the long composite time range (see figure \ref{fig:sample_output}). This highlights the importance of multi-sensor fusion where Sentinel-1 backscatter provides valuable information for these gaps since SAR is not affected by clouds. In order to reduce the noise in SAR backscatter signal, we apply the same compositing method to the VV and VH bands without the cloud mask. In section \ref{sec:results} we will discuss in detail the model performance in challenging regions with imperfect cloud free composites.

\subsubsection{Global dataset}
\label{sec:global_dataset}
For each sampled tile, we generate Sentinel-1 and Sentinel-2 composites according to the definition in section \ref{sec:cloud_mask} by gathering imagery using the Descartes Labs platform for the year 2021 (2021/2022 for the southern hemisphere). The DEM has a fixed acquire year of 2000. For a given tile, we gather all GEDI level-2a/b and level-4a point data which lay within the tile geometry and have a collection date no more than $\pm$1 month from the composite time range (this buffer is set to 0 months for the 6 months composites). In order to match the data from GEDI level-2a/b with level-4a, they are required to have the same footprint coordinates to 5 decimal point precision and the same acquire date. Further, we only accept data with quality flag \textit{l4\_quality\_flag} equal to 1 and require at least 20 data points to be within the tile geometry. Due to the sparsity of the GEDI data, it is saved as vector data along with each footprints pixel coordinate and rasterized on the fly during training. Despite the footprint being 25~m in diameter, we assign the corresponding target value to only one pixel of size 10~m$\times$10~m at the location of the footprint center.
\\ \\
In total 276'766 data samples were created of which 14'745 (5\%) are randomly selected and stored as test dataset. Each sample is composed of an average of 50 scenes across Sentinel-1 and Sentinel-2 amounting to a total of 13.8~M scenes processed. The total number of ground truth data points is 67~M (3.8~M) for the training (test) dataset.

\section{Methodology}
\label{sec:methodology}
In this section we describe our approach to building a computer vision (CV) model which fuses selected bands of multiple sensors as well as encoded geographic location to form a 13 channel input tensor and simultaneously predicts a continuous map of above ground biomass density, canopy height, canopy cover and their respective uncertainties at the same resolution as the input imagery.

\subsection{Model development}
\label{sec:model_development}
\begin{figure}[h]
	\centering
	\includegraphics[width=\textwidth]{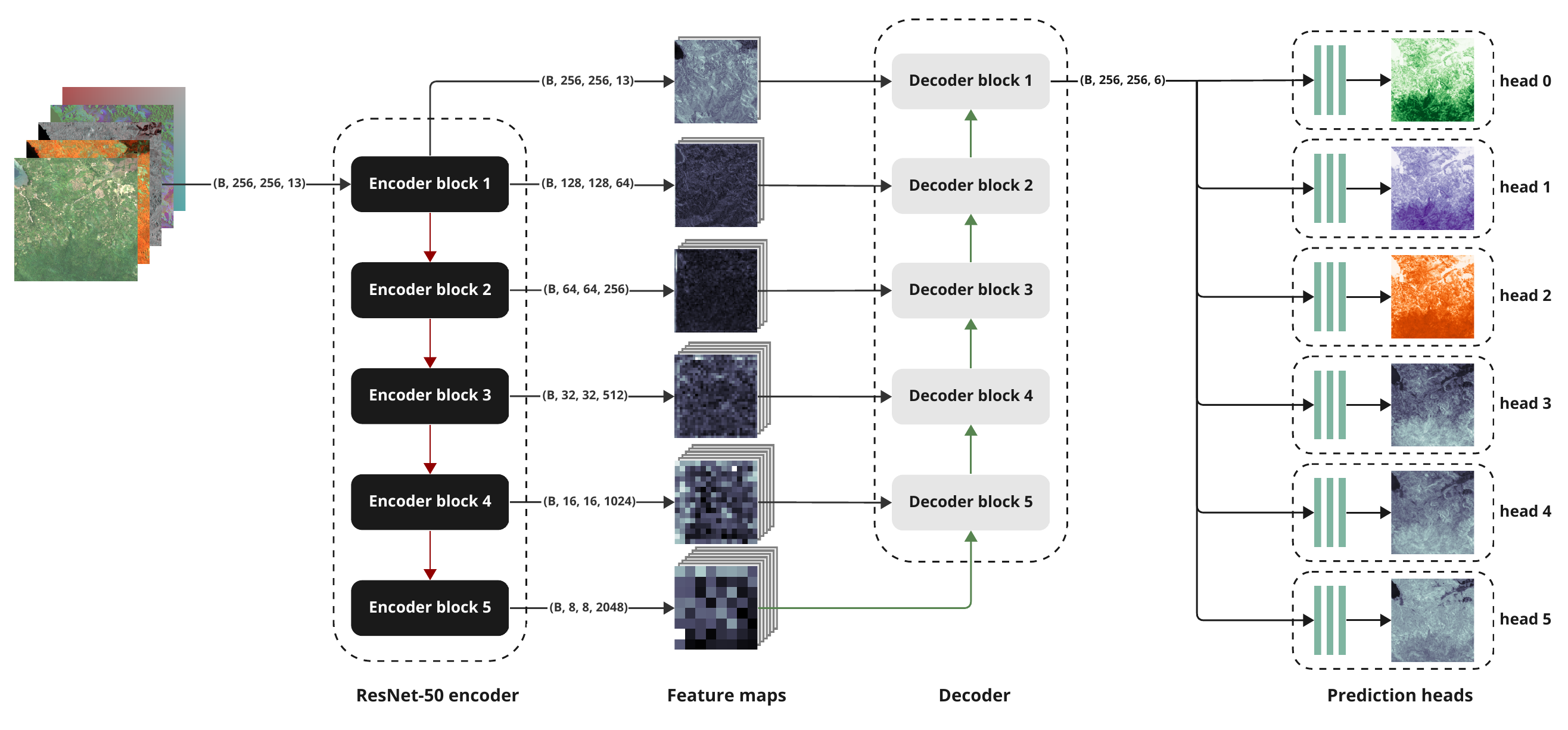}
	\caption{Simplified version of the model architecture}
	\label{fig:model_architecture}
\end{figure}

The GEDI dataset offers measurements of AGBD, CH and CC for recent years but is largely incomplete at higher resolution due to its sparsity. Here we describe the model we developed which uses multiple source as input to predict AGBD, CH and CC as a continuous map at the resolution of the input source while using the GEDI level-2 and level-4 dataset for training.
\begin{figure}[h]
	\centering
	\includegraphics[width=0.8\textwidth]{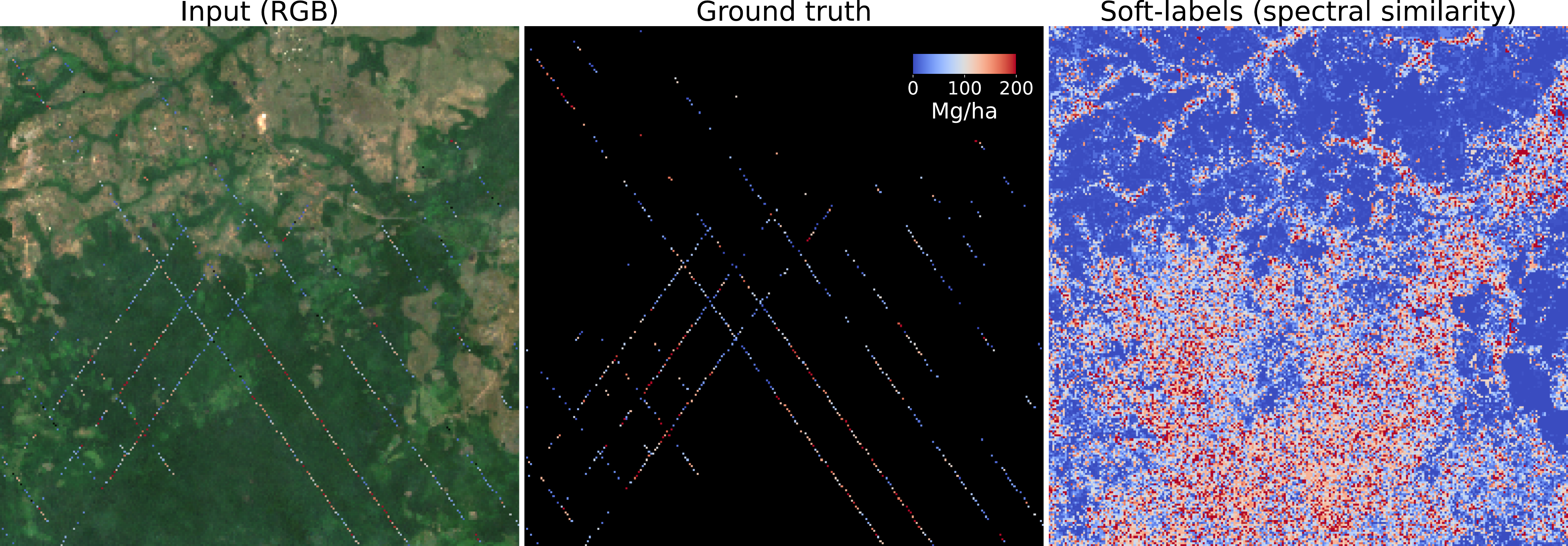}
	\caption{Sample input image in RGB with overlaid AGBD ground truth measurements (left), rasterized AGBD GEDI footprints (middle) and soft labels (right)}
	\label{fig:input_sample}
\end{figure}
In this work, we use the surface level processed Sentinel-2 with 6 bands (red, green, blue, nir, swir1, swir2), Sentinel-1 backscatter signal (bands VV and VH) as well as altitude, aspect and slope provided by the digital elevation model from the SRTM mission at a resolution of 10~m. The choice of Sentinel as input source is motivated by its global coverage, the relatively high spatial and temporal resolution and the fact that the image collection goes back to 2016 allowing the deployment of the model on historical data.
\\ \\
In figure \ref{fig:input_sample} a sample input image (RGB) and the corresponding AGBD ground truth data is shown. It illustrates the sparsity of the ground truth measurements and the need for a model which can fill in the gaps. Our model consists of a convolutional neural network (CNN) (\cite{lecun1998gradient}) with three main components: an encoder network which extracts features from the input image, a decoder which processes the extracted feature maps at different depths of the network and, together with the encoder, forms a feature pyramid network (FPN). The final feature map then consists of all relevant information required for the estimation of the output variables. The final components are a set of prediction heads which have the function of generating the estimate of each respective output variable based on the last feature map of the FPN. Each prediction head consists of a series of 1$\times$1 convolutions. The encoder network can be any commonly used feature extractor. In this work we chose ResNet-50 (\cite{he2016deep}) as the encoder. The FPN consists of decoder blocks where each block takes the features from level $l-1$ and $l$ as input. The feature map of level $l-1$ is up-sampled using a bilinear interpolation method and a convolution layer with kernel size 2$\times$2, before concatenating with the feature map of level $l$ followed by two convolution layers with kernel size 3$\times$3. The resulting feature map is then fed to the decoder block of level $l+1$ until reaching the final level corresponding to the input resolution. All convolution layers in the decoder have a fixed feature dimension of 128. Each prediction head consists of a series of 3 1$\times$1 convolutions with feature dimensions [128, 128, 1]. All layers in the decoder and prediction heads use the ReLU (\cite{agarap2019deeplearningusingrectified}) activation function except for the final layer of the prediction heads which estimate the variable uncertainty for which we use the softplus (\cite{7280459}) activation function. The weights of the full network are randomly initialized using the Glorot Uniform (\cite{pmlr-v9-glorot10a}) initializer. In figure \ref{fig:model_architecture} a simplified version of the model architecture is shown. The total number of trainable parameters is 28.71 million.

This model architecture generates an output map of the same size as the input image. In most cases of such a configuration, the target is expected to be a continuous map of the same size as the output from which the loss is calculated. In our case, the target is not continuous but is composed of sparse target values. A straightforward solution for this situation is to only evaluate the loss function at pixels for which a ground truth value is available. However, we found that this approach results in over-fitting on the sparse pixels and generates a non-homogeneous output. We therefore propose a new approach where we generate a continuous target map by a model acting as the teacher to a student network during the training process. This approach is similar to the student-teacher setup for classification tasks where ground truth labels are incomplete. We extend this approach to segmentation tasks such as in this work. To begin with, we construct two identical networks in terms of the architecture whose weights are initialized separately and referred to as teacher ($\mathcal{F}_{T}$) and student ($\mathcal{F}_{S}$) networks. The task of $\mathcal{F}_{T}$ is to generate a continuous ground truth map from the input and the ground truth labels. $\mathcal{F}_{S}$ is then trained on this ground truth map. After a certain amount of epochs, $\mathcal{F}_{T}$ and $\mathcal{F}_{S}$ swap their roles. This procedure is repeated until the two networks converge. Since both networks are randomly initialized, they are not very skillful during the initial training phase and $\mathcal{F}_{T}$ can not provide meaningful ground truth guesses. We therefore replace $\mathcal{F}_{T}$ with a simple model based on the spectral similarity at the input level during the first training phase of $\mathcal{F}_{S}$. We define the spectral similarity as the cosine similarity
\begin{equation}
    \label{eq:spectral_similarity}
    \sigma\left(X_{i},X_{j}\right)=\frac{X_{i}\cdot X_{j}}{\| X_{i} \| \: \| X_{j} \|}
\end{equation}
where $X_{i}$ and $X_{j}$ are the vectors with spectral information of a set of bands for pixel $i$ and $j$. The set of bands can be a combination of any of the available input bands. However, we find that this approach works best with a feature vector composed of all 6 bands from Sentinel-2.
\begin{figure}[h]
	\centering
	\includegraphics[width=0.8\textwidth]{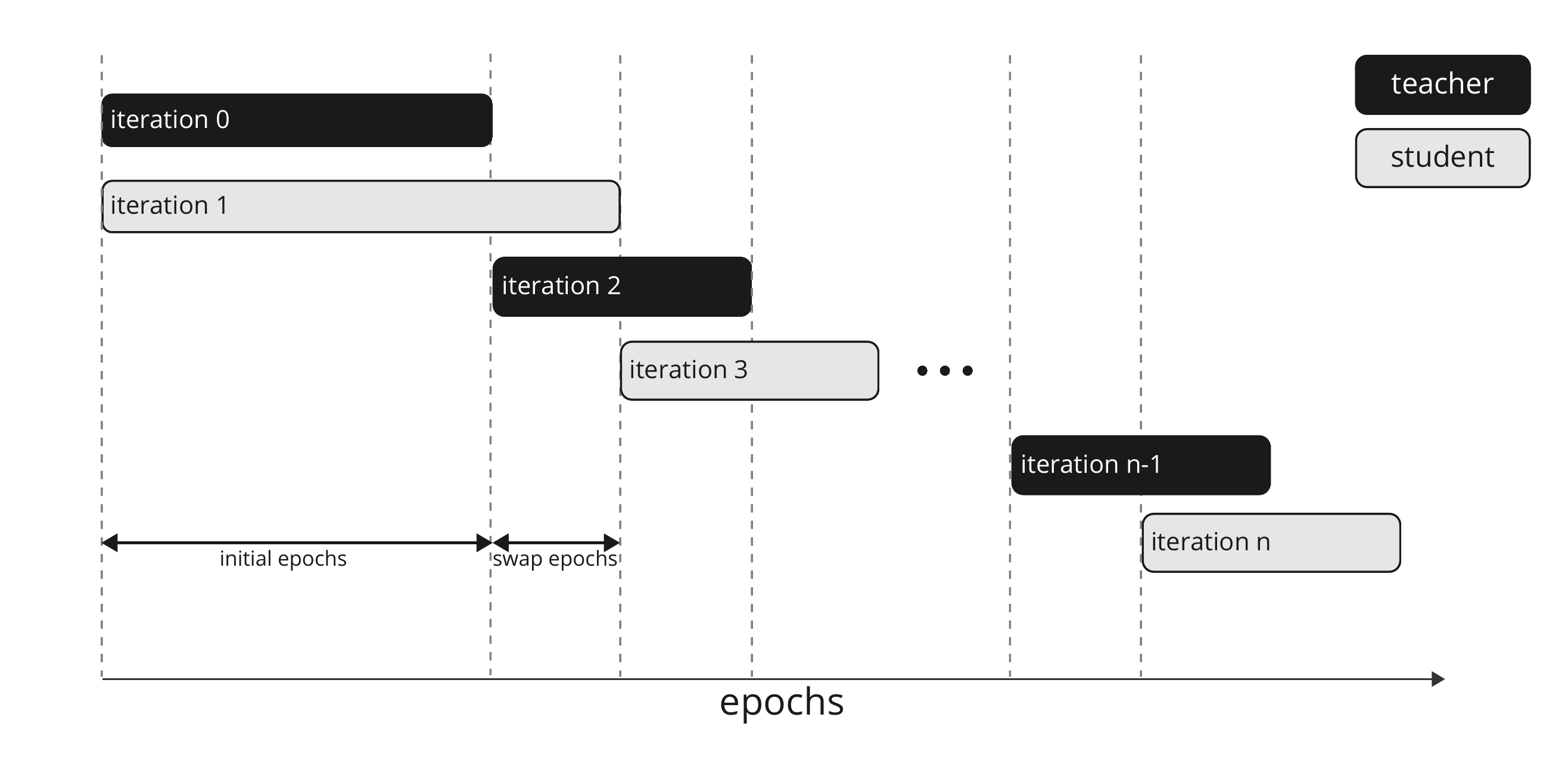}
	\caption{Illustration of the training process with two identical networks acting as teacher and student models which periodically switch their roles.}
	\label{fig:student_teacher}
\end{figure}
For each pixel with no label, we compute $\sigma\left(X_{i},X_{j}\right)$ with respect to all pixels which have a label (hard labels). We then assign the value of the hard label for which $\sigma\left(X_{i},X_{j}\right)$ is maximal to the pixel with no label to generate a soft label. The target map is then a combination of hard and soft labels defined as
\begin{equation}
    \label{eq:softlabels_and_hardlables}
    \widehat{y}=m \otimes \widehat{y_{h}} + (1-m) \otimes \widehat{y_{s}}
\end{equation}
where $m$ is a mask for which $m_{i,j}=1$ if pixel $i,j$ is a hard label and $m_{i,j}=0$ otherwise. Here $\otimes$ denotes the element wise product. The soft labels can be calculated efficiently for all non-labeled pixels by arranging them in a matrix $A$ of size $n \times b$ and all hard label pixels in a matrix $B$ of size $m \times b$ with $b$ the number of bands. The soft labels are then calculated by
\begin{equation}
    \label{eq:softlabels}
    \mbox{label}_{idx}=\mbox{argmax}\left(\widehat{A}\widehat{B}^{T}\right)
\end{equation}
where $\widehat{A}$ and $\widehat{B}$ are the row normalized matrices of $A$ and $B$. Figure \ref{fig:input_sample} (right) shows the result of this operation on the input shown in figure \ref{fig:input_sample} (left).\\ \\
Here we use a simple model to generate soft labels according to eq. \ref{eq:spectral_similarity} which are good prior guesses for pixels with no target values in the first iteration of training $\mathcal{F}_{S}$. After the first iteration (initial epochs), $\mathcal{F}_{S}$ has acquired some skills to predict the target map and becomes the teacher to the second network which is trained from scratch. In order for the new student network to overcome the skills of the teacher network, the hard labels will be given more weight than the soft labels in the definition of the loss function (see eq. \ref{eq:balanced_loss}) and $\mathcal{F}_{S}$ is trained for 8 more epochs (swap epochs) in the current iteration than $\mathcal{F}_{T}$ in the previous iteration. This procedure is illustrated in figure \ref{fig:student_teacher}. The soft labels generated by $\mathcal{F}_{T}$ are guesses and help guide the network training, particularly in the early training phase but may deviate from the actual ground truth label. We therefore use a weight schedule for the soft labels incorporated into the loss function (see the next section for details).

\subsection{Model training}
\label{sec:model_training}
Training of the model is divided into three parts: First we train the entire network including prediction heads 1-3 on the full dataset of $\sim$1~M samples. The network is optimized to simultaneously predict AGBD, CH and CC using their respective target values. We incorporate sample weighting according to the inverse probability distribution function (PDF) of the respective variable distributions. This is important in order to mitigate over fitting on lower values of AGBD and CH as they appear at higher frequency in the dataset. In the second stage we freeze the weights of the encoder and decoder and fine tune the prediction heads 1-3 separately on variable specific datasets. These datasets are subsets of the original dataset with a more uniform distribution of variable specific values. This is done by excluding a certain number of samples according to their aggregated target values which is further described in appendix \ref{apx:uniform_sampling}. The balanced datasets still manifest some non-uniformity in the distribution of individual point measurements, as opposed to aggregated measurements within a sample. We therefore incorporate an adjusted sample weighting according to the inverse PDF of the respective variable distributions in the uniform dataset. In the third and last stage we fine tune the pairs of prediction heads [(1, 4), (2, 5), (3, 6)], i.e. each variable and its uncertainty, separately on the same datasets as in stage 2.

\subsubsection{Loss function}
\label{sec:loss_function}
We consider the prediction of each pixel $(i,j)$ in the output map $y_{i,j}$ as an independent measurement of a normal distributed variable with a standard deviation of $\sigma_{i,j}$. The probability for a given ground truth value $\hat{y}_{i,j}$ is then given by
\begin{equation}
    \label{eq:probability_distribution}
    p(\hat{y}_{i,j}|y_{i,j}(\theta), \sigma_{i,j}(\theta))=\frac{1}{\sqrt{2\pi\sigma_{i,j}^{2}(\theta)}}\mathrm{e}^{-\frac{\left(\hat{y}_{i,j}-y_{i,j}(\theta)\right)^{2}}{2\sigma_{i,j}^{2}(\theta)}}
\end{equation}
where $\theta$ are the parameters of the network. The likelihood function can be written as
\begin{equation}
    \label{eq:likelihood_function}
    L(\theta)=\prod_{i,j}p\left(\hat{y}_{i,j}|y_{i,j}(\theta),\sigma_{i,j}(\theta)\right).
\end{equation}
We use gradient descent to optimize the parameters $\theta$ which minimize the negative log likelihood (NLL) which defines our loss function
\begin{equation}
    \label{eq:nll}
    \mathcal{L}=-\log(L)=\frac{1}{2}\sum_{i,j}\left(\frac{\left(\hat{y}_{i,j}-y_{i,j}(\theta)\right)^{2}}{\sigma_{i,j}^{2}(\theta)}+\log(\sigma_{i,j}^{2}(\theta))\right)
\end{equation}
where we omitted the factor $2\pi$ from equation \ref{eq:probability_distribution}. $\sigma_{i,j}(\theta)$ is the uncertainty predicted by the model for each variable separately. By definition, we expect 68~\% of all samples to have an absolute error between predicted and true value within the range of 1$\sigma$. During training we verify this by calculating the fraction of z-scores, defined as $z=|\hat{y}-y|/\sigma$, to be <1. Even though the $\log(\sigma^{2})$ term in equation \ref{eq:nll} acts as regularization to make sure the model does not learn a trivial solution by predicting a very large $\sigma$, we noticed that the coverage may still be >0.68. We therefore introduce an addition regularization term in the definition of the loss as
\begin{equation}
    \label{eq:loss_regularized}
    \mathcal{L}=-\log(L)+\lambda\sigma^{2}
\end{equation}
where $\lambda$ is a hyper parameter determined for each variable separately. For a given sample, the number of hard labels is much smaller than the number of soft labels (on average the ratio of hard to soft labels is $\sim$1/1000) and varies between samples. We therefore introduce a pixel weighting scheme to balance the contribution of hard and soft labels to the loss. In addition to the imbalance between hard and soft labels, we also want to assign relative weighting of hard to soft labels. This is an essential requirement in order to make the student-teacher approach work well. Let us consider the relative weight of a hard label to be $\lambda_{h}$ and that of a soft label to be $\lambda_{s}$. Then the balanced loss function, taking both the relative weights as well as the number of hard and soft label pixels into account, becomes
\begin{equation}
    \label{eq:balanced_loss}
    \mathcal{L}_{b}=\left(\frac{\lambda_{h}}{n_{h}}m+\frac{\lambda_{s}}{n_{s}}(1-m)\right)\otimes\mathcal{L}
\end{equation}
where $n_{h}$ ($n_{s}$) are the number of hard (soft) label pixels in a given sample and $m$ has the same definition as in equation \ref{eq:softlabels_and_hardlables}. By default, we choose $\lambda_{h}=1$ and vary $\lambda_{s}$ during training according to an exponential decay from 1 to 1e-3 during the initial epochs, then exponentially increase it to 1e-2 during the remaining epochs. We have considered other schedules such as linear, constant and zero (corresponding to no soft label) which all resulted in worse model performance.
\\ \\
So far we have only formulated the loss function in equation \ref{eq:balanced_loss} considering one variable. However, we train the model for all variables simultaneously for which we construct the final loss as the weighted sum over the variable specific components
\begin{equation}
    \label{eq:final_loss}
    \mathcal{L}_{b}^{\mathrm{total}}=\alpha_{0}\mathcal{L}_{b}^{AGBD}+\alpha_{1}\mathcal{L}_{b}^{CH}+\alpha_{2}\mathcal{L}_{b}^{CC}
\end{equation}
where the weights $\alpha_{i}$ allow for balancing the different contributions due to the different target scales of the variables. In this work we set $\alpha_{0}=\alpha_{1}=\alpha_{2}=1$ since we did not observe any improvements using variable specific weighting.

\subsubsection{Training details}
\label{sec:training_details}
For the first stage of pre-training on the full global dataset, we train the model for 40 epochs with a batch size of 72 on a multi-GPU node with 4 A10G GPUs. We reserve one of the GPUs for data pre-processing, such as calculation of the soft labels, while the remaining GPUs perform the model training. We use the Adam optimizer (\cite{kingma2017adammethodstochasticoptimization}) with a linearly increased learning rate from 1e-7 to 1e-4 over a warm up period of 1 epoch after which it is continuously decreased according to a cosine function over the remaining training period. The second stage consists of fine tuning each variable separately on the balanced datasets and applying the sample weighting according to the inverse frequency distribution. In this stage only the prediction heads 1-3 are trained while keeping all other model weights frozen. We use three single GPU nodes to train each head in parallel with a batch size of 32. In the third and last stage, we fine tune the prediction heads 4-6 which are responsible for predicting the variable uncertainty. In all stages, the loss function as defined in equation \ref{eq:final_loss} is minimized. However, in stages one and two the uncertainty estimation is ignored which is equivalent to setting $\sigma=1$ for all pixels.

\section{Results}
\label{sec:results}
In this section we discuss the results of evaluating our model both qualitatively and quantitatively on the test dataset as well as independent datasets which focus on one of the three variables which our model predicts. We first show some qualitative results by deploying the model over specific regions covering a diverse range of biomes. We then evaluate the model using the held out global test dataset and determine a number of metrics such as mean error (ME), mean absolute error (MAE), mean absolute percentage error (MAPE), root mean squared error (RMSE) and correlation for each of the predicted variables separately. Finally, we determine our models accuracy against publicly available datasets.

\subsection{Qualitative assessment}
\label{sec:qualitative_assessment}
The deployment of the model at a global scale for the year 2023 allows for a qualitative assessment of its performance, in particular in challenging areas such as those with high cloud coverage which may affect the quality of the input imagery.
\begin{figure}[h]
	\centering
	\includegraphics[width=\textwidth]{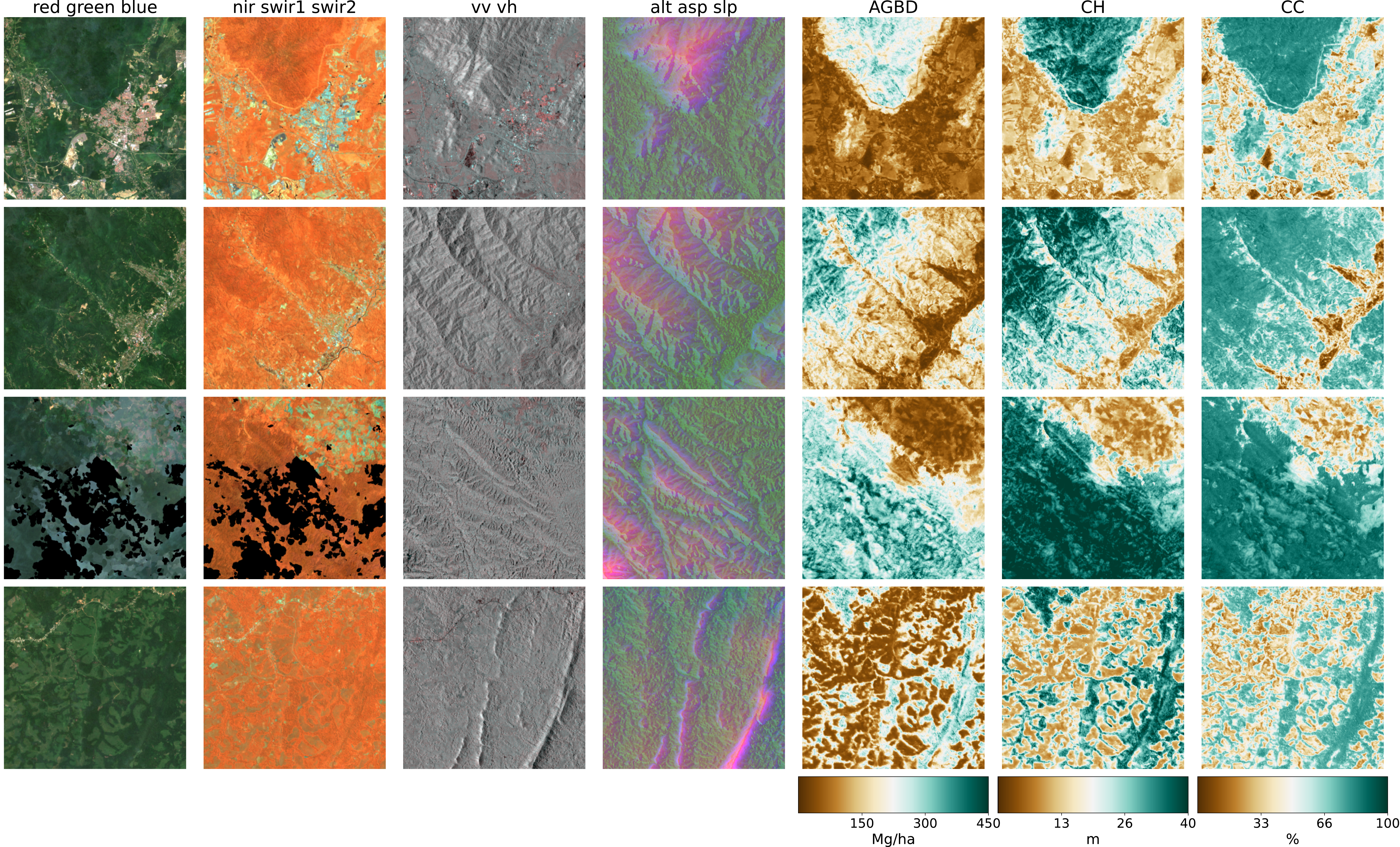}
	\caption{Qualitative assessment of the model performance on four sample locations. The first four columns from the left illustrate the respective input data while the three columns form the right corresponds to the models predictions.}
	\label{fig:sample_output}
\end{figure}
In figure \ref{fig:sample_output} we present four sample locations which all cover an area of 1~km$\times$1~km at 10~m resolution. The left four columns represent the input data (excluding the geographic coordinates) while the right three columns correspond to the predictions of AGBD, CH, CC. The top row depicts a sample where the model performs very well. It contains various land-cover types such as urban, agriculture and low- as well as high-density forest areas. The second row shows a sample in high-density mountainous area with a non-forested valley. Both SAR backscatter and the DEM provide valuable information on the terrain in this sample. The model performs as expected and successfully distinguishes high- from low-density and forested from non-forested areas. The third row demonstrates the models ability to leverage the multi-sensor input stack. There are multiple regions where the cloud free composite left gaps visible as black areas in Sentinel-2 data. Here, information from Sentinel-1 and DEM is used to enhance the predictability for these areas. The last row illustrates the models capability for making accurate predictions at high resolution. The example shows individual trees, or groups of trees, which are well separated from the bare ground.

\subsection{Quantitative assessment}
\label{sec:quantitative_assessment}
\begin{figure}[h]
	\centering
	\includegraphics[width=\textwidth]{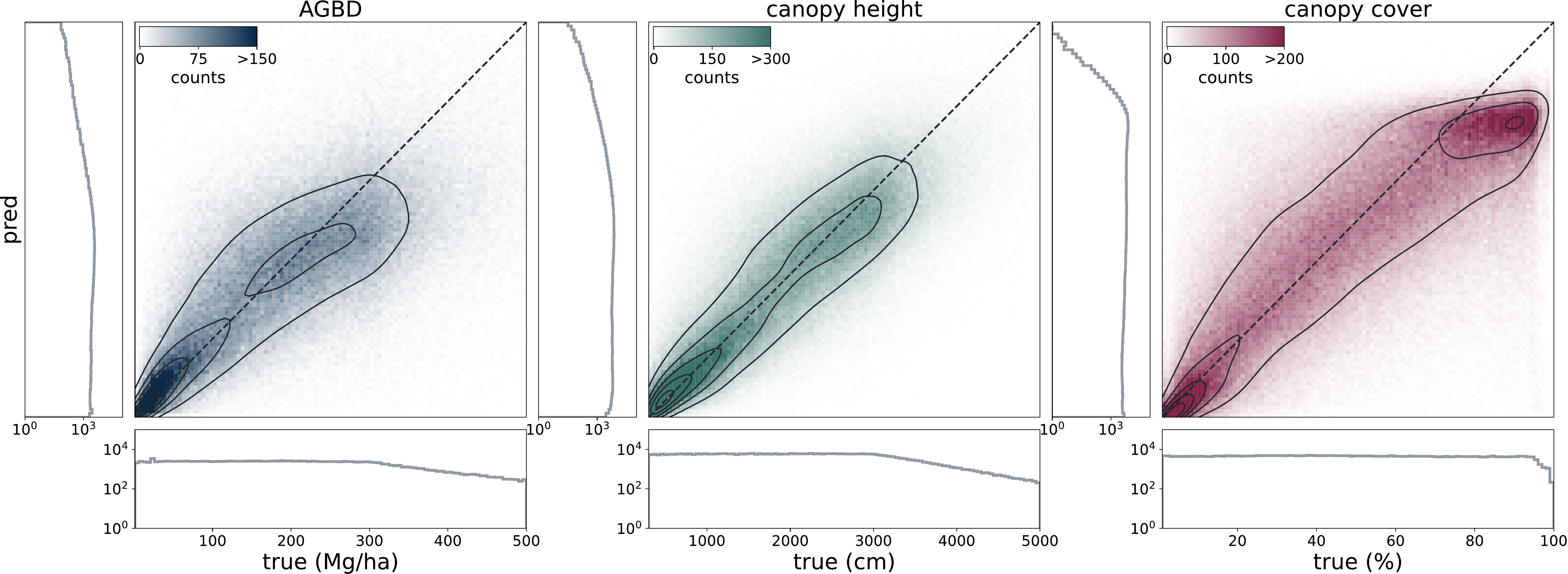}
	\caption{Predicted vs. true values for all samples in the test set at the footprint level after applying inverse frequency sampling for a uniform distribution.}
	\label{fig:y_pred_y_true}
\end{figure}

\begin{figure}[h]
	\centering
	\includegraphics[width=\textwidth]{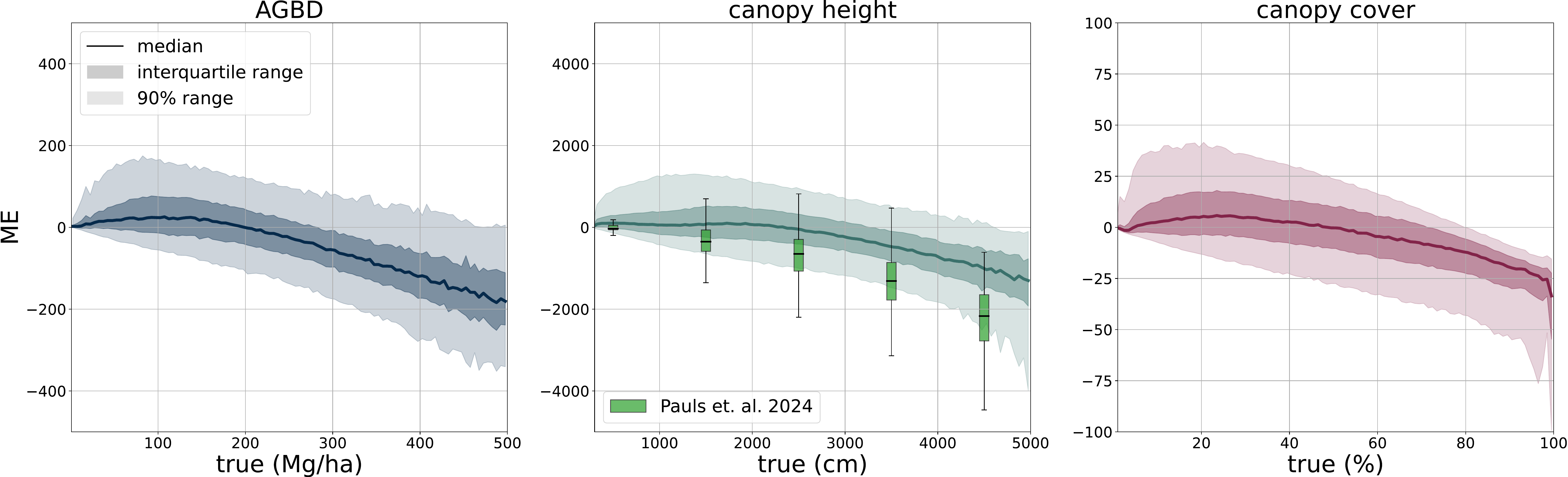}
	\caption{Median, 50~\% and 90~\% range of errors between predicted and true values in each bin vs true value for the variables AGBD (left), CH (middle) and CC (right).}
	\label{fig:me_binned}
\end{figure}

\begin{figure}[h]
	\centering
	\includegraphics[width=\textwidth]{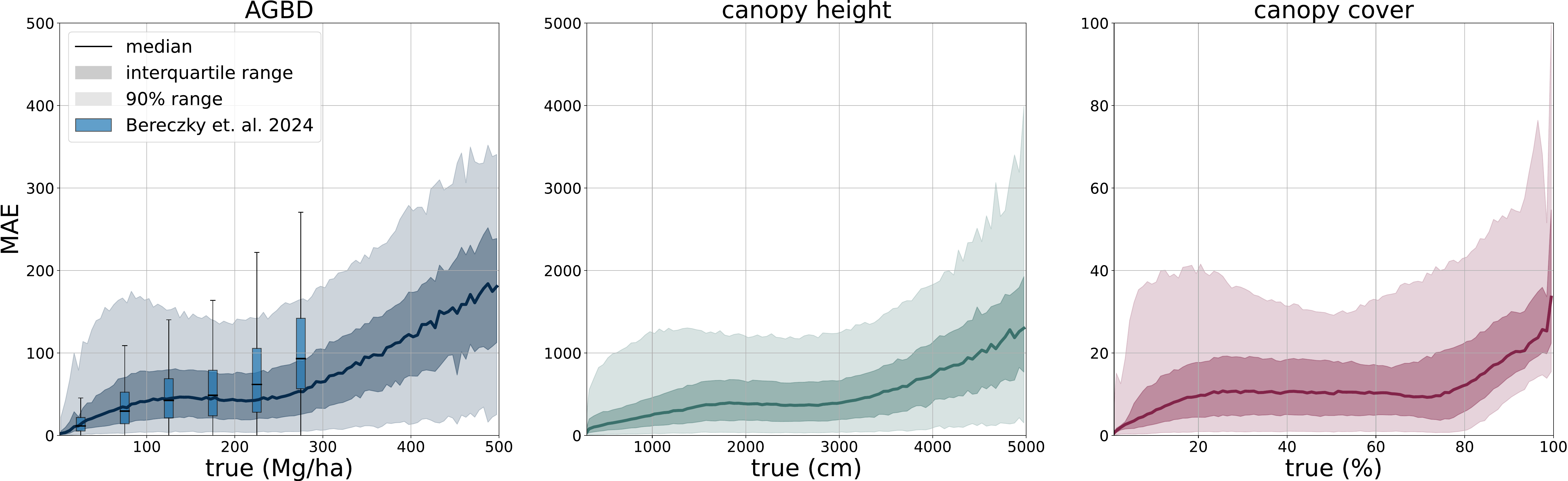}
	\caption{Median, 50~\% and 90~\% range of absolute errors between predicted and true values in each bin vs true value for the variables AGBD (left), CH (middle) and CC (right).}
	\label{fig:mae_binned}
\end{figure}

\begin{figure}[h]
	\centering
	\includegraphics[width=\textwidth]{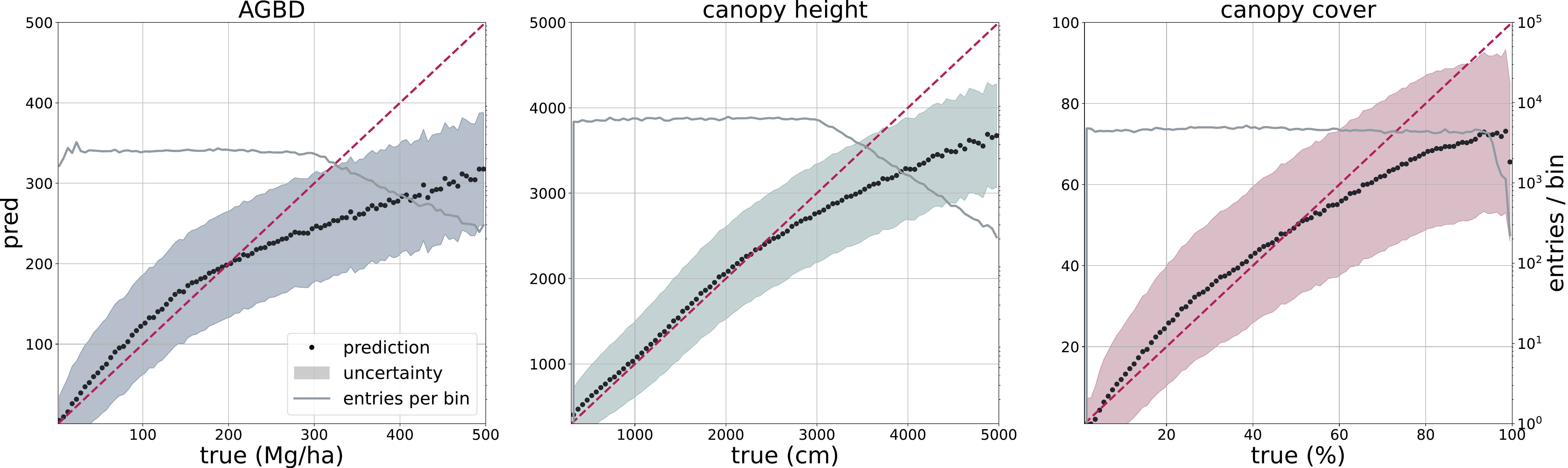}
	\caption{Median of predicted values and 1$\sigma$ uncertainty range within each bin as a function of the true values.}
	\label{fig:y_pred_y_true_median}
\end{figure}

We have evaluated the performance of our model based on the test dataset consisting of 14'745 samples and 13.8~M individual GEDI footprints. Model inference is performed on each test sample of size 256~pixels$\times$256~pixels which generates predictions for every pixel and all output variables including their uncertainties. We measure the correlation, mean error (ME), mean absolute error (MAE), mean absolute percentage error (MAPE) and root mean squared error (RMSE) between the GEDI ground truth data and the model predictions gathered at the pixel coordinates of the GEDI footprints. Due to the non-uniform distribution of ground truth values, we assess the model performance on both the test set with its original as well as a uniform sample distribution. For this purpose, we sample data points according to the inverse PDF of each variables respective distribution. This ensures a more fair performance assessment across the value ranges of each variable. Due to the low frequency of occurrence of high values of AGBD and CH, we set a reference value of 300~Mg/ha for AGBD and 30~m for CH to define the lowest value of probability. Values with lower probabilities are always included during sampling. After the sampling procedure, a total number of 200~k - 450~k data points are remaining, depending on the variable. In table \ref{tab:result_summary} we summarize the evaluation metrics.
\begin{table}[h!]
\centering
\begin{tabular}{c|c|c|c|c|c|c|c|c|c|c}
\hline
 & \multicolumn{5}{|c|}{uniform} & \multicolumn{5}{|c}{original} \\
variable & corr & ME & MAE & MAPE & RMSE & corr & ME & MAE & MAPE & RMSE \\
\hline
AGBD & 0.70 & -10.83 & 62.95 & 48.03 & 86.74 & 0.83 & 10.75 & 26.07 & 166.60 & 50.59 \\
CH & 0.80 & -20.09 & 455.46 & 27.56 & 624.94 & 0.85 & 83.51 & 370.95 & 35.19 & 543.81 \\
CC & 0.75 & -3.40 & 13.92 & 45.26 & 18.68 & 0.88 & -1.92 & 9.86 & 104.10 & 15.75 \\
\hline \hline
\end{tabular}
\caption{Summary of evaluation metrics for all prediction variables and both the original test set and a subset created by uniform sampling across all variables. The units for AGBD (CH, CC) are Mg/ha (cm, \%) except for correlation which is unit-less and MAPE which is given in \%.}
\label{tab:result_summary}
\end{table}
These metrics are representative of the model performance across the full range of variable values. In order to get a better understanding of the model performance at various ground truth values, we also create binned evaluation metrics. We aggregate the sample pairs $\mathbf{y}^{i}=(y_{\mbox{true}}^{i}, y_{\mbox{pred}}^{i})$ for each variable $i \in {0, 1, 2}$ where $y_{\mbox{true}}^{i}$ falls within a specific bin and calculate the metrics for each bin separately. The bin range and bin size $(b_{\mbox{low}}, b_{\mbox{high}}, b_{\mbox{size}})$ for AGBD is (0, 500, 5)~MG/ha, for CH (0, 5000, 50)~cm and CC (0, 100, 1)~\%. Figure \ref{fig:y_pred_y_true} shows a scatter plot of samples $\mathbf{y}^{i}$ for each variable $i$ where the color map represents the sample density. The distributions of $y_{\mbox{true}}$ and $y_{\mbox{pred}}$ are shown in the bottom and left panels. Overall, there is very good agreement between predicted and true value across the full variable ranges. However, there is an increased disagreement at higher values of AGBD and CH. This can both be attributed to the lower amount of training samples in these regions as well as a signal saturation in the input data.
In figure \ref{fig:me_binned} (\ref{fig:mae_binned}) the median error (absolute error) within each bin (solid line) as well as interquartile range\footnote{Interquartile range corresponds to the range which contains 50~\% of all data points.} (dark shaded area) and 90~\% range (light shaded area) is shown. The plots further illustrate the good agreement of the models predictions with ground truth data. For AGBD we observe a slight over-prediction within the range 0-200~Mg/ha and a slight under-prediction for values >200~Mg/ha. The CH predictions agree very well up to values of 25~m where the model starts to slightly under-predict. For CC the agreement is very good across the range of values except for full coverage (close to 100~\%) where the model tends to slightly under-estimate. It is worth noting that evaluation metrics determined across the full variable range (as shown in table \ref{tab:result_summary}) are biased towards larger values due to outliers which can have a large contribution when using a uniformly sampled dataset. On the other hand, the metrics may be biased towards smaller values when using the original sample distribution because lower values (and therefore smaller errors) are more frequent. This is an important consideration when comparing results across different works since the sample distributions, and the ranges within they are defined, are different. In this work, we present all metric plots based on the uniformly sampled dataset, as this provides the least bias, and include the results on the original dataset in table \ref{tab:result_summary}.
We compare our results to the most recent and state-of-the-art results from (\cite{Bereczky2024}) for AGBD and (\cite{pauls2024estimatingcanopyheightscale}) for CH. Their corresponding error metrics are included in figure \ref{fig:mae_binned} (left) and figure \ref{fig:me_binned} (middle). Our model performance on AGBD is comparable with (\cite{Bereczky2024}) in the range of 0-200~Mg/ha but shows improved performance above 200~Mg/ha which corresponds to regions where a large portion of the worlds biomass is stored, such as rain forests, and is therefore of great importance. For CH, our model outperforms (\cite{pauls2024estimatingcanopyheightscale}) across the entire range except for small trees (<10~m) where both approaches show comparable results. We did not find previous work on CC that we can compare our results to.
\\ \\
Finally, figure \ref{fig:y_pred_y_true_median} shows the median predictions within each bin as a function of the true values. The shaded areas correspond to the median standard error prediction of all samples within a given bin. These plots further demonstrate the good agreement of our models predictions with ground truth data in the low to mid variable range while it tends to under-estimate at high values of the prediction variables. The under-estimation for AGBD and CH can be attributed to the lack of training data at very high values as well as a saturation effect at the input level. For AGBD, there is an additional effect arising from increasing uncertainties of the ground truth at higher values of AGBD. We further discuss the implications on the prediction uncertainties in appendix \ref{apx:gedi_uncertainty}. The 1$\sigma$ uncertainty bands are slightly over-estimated at lower values for all variables as the coverage exceeds 68~\% and is slightly under-estimated at higher values where the coverage is <68~\%. We reserve addressing this discrepancy for future work but provide the exact coverages in appendix \ref{apx:uncertainty_estimation}.
\\ \\
The above model performance assessment is conducted using samples across all world regions and plant functional types (PFT). To further assess the models accuracy with respect to the different PFTs, we repeat the analysis on samples grouped by PFT where we retrieve the PFT class from the GEDI dataset and are categorized as: Deciduous Broadleaf Trees (DBT), Evergreen Broadleaf Trees (EBT), Evergreen Needleleaf Trees (ENT) and Grasslands/Shrublands/Woodlands (GSW). The results of this analysis are summarized in appendix \ref{apx:assessment_pft}. Overall, the model performs best on samples of class ENT followed by DBT and EBT. The errors are lowest for class DSW, however, this is due to the fact that most samples are at very small values for which the errors are small on an absolute scale.

\subsection{Assessment against third party datasets}
\label{sec:third_party_datasets}
In the following section we present the results from studies where we compared our models output with third party datasets that are independent of the GEDI measurements. All these datasets were either generated from high resolution, airborne LiDAR or on-the-ground measurements. We deployed our model over the respective regions for the years the ground truth data was collected. We divide our studies into estimation of above ground biomass and canopy height since the dataset are focused on one of these variables.

\subsubsection{Above ground biomass}
\label{sec:third_party_biomass}
\begin{figure}[h]
	\centering
	\includegraphics[width=0.8\textwidth]{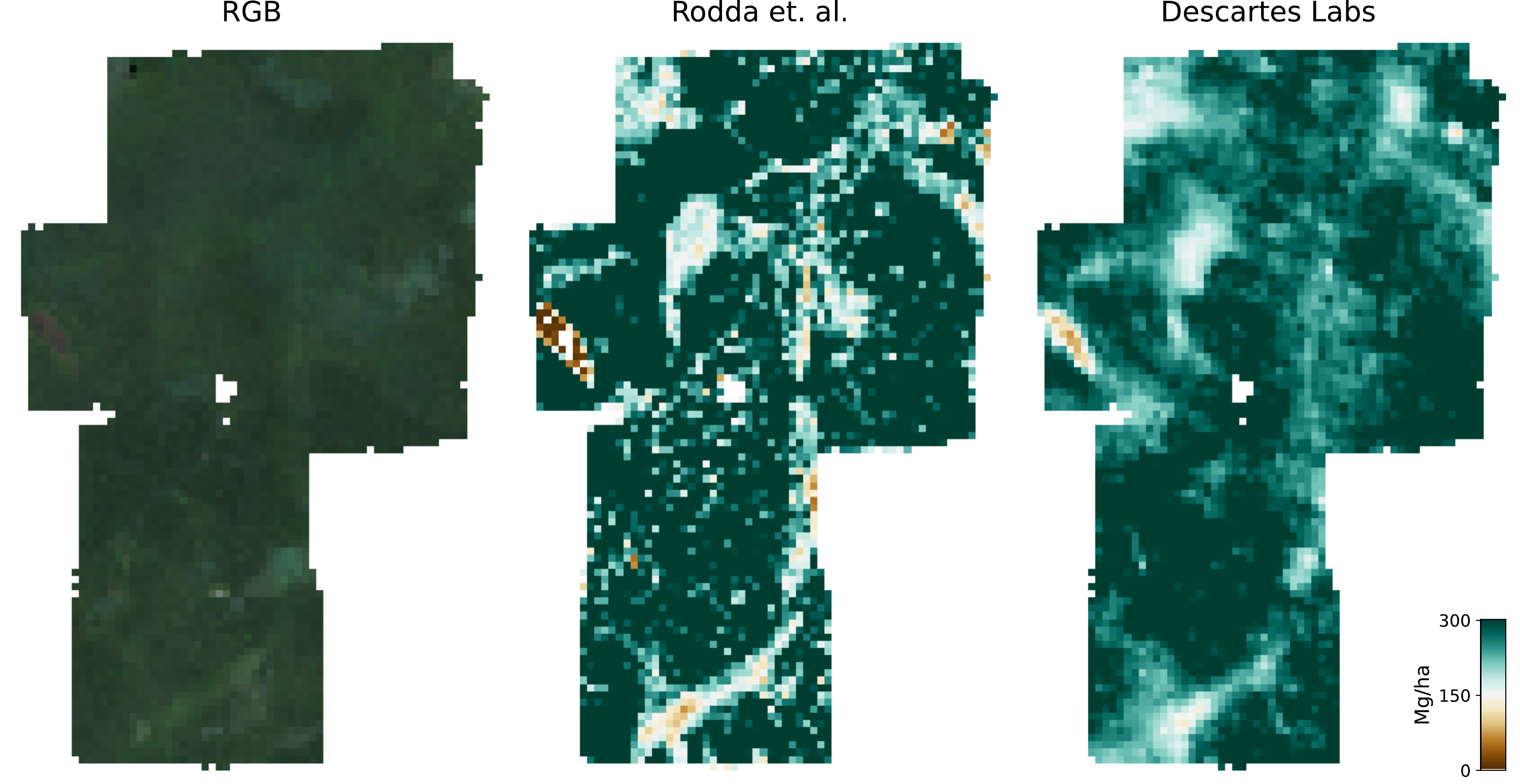}
	\caption{Sample data from a plot in Central Africa. Left: RGB image of the input composite. Middle: Ground measurement from Rodda et. al. Right: Estimate of aboveground biomass density from our model.}
	\label{fig:isro_samples}
\end{figure}

\begin{figure}[h]
	\centering
	\includegraphics[width=\textwidth]{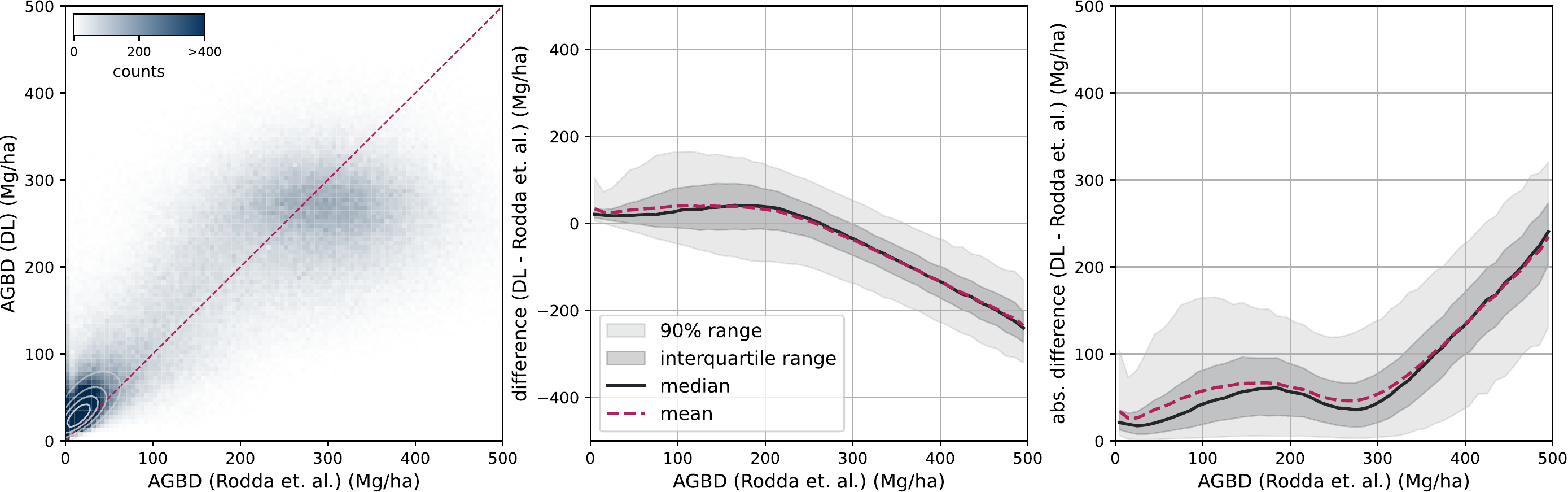}
	\caption{Histogram of estimated AGBD in each 40~m$\times$40~m pixel by our model vs. the ground measurements by Rodda et. al. (left), median (middle) and absolute median (right) error as well as means, interquartile and 90~\% ranges in bins of size 5~Mg/ha with respect to AGBD by Rodda et. al.}
	\label{fig:isro_results}
\end{figure}

For the verification of biomass estimates, we utilize the dataset created by (\cite{RODDA2024}) which consists of 13 plots of variable size at a maximum resolution of 40~m. 8 of the plots are located in the Central Africa region and 5 of the plots in the South Asia region. This dataset was compiled from forest inventory collected within the respective sites during different time ranges. Site-level measurements followed a strict protocol where the diameter at breast height (DBH) was determined for each individual tree within the plots as well as the tree height (H) for a subset of trees. The tree level taxonomic identification and relative coordinates within the plots were recorded along with geographic coordinates of the plot borders at intervals of 20~m. The forest inventories were split into 1~ha and 0.16~ha plots. The collected data of DBH-H on a subset of trees within these plots was used to fit allometric equations which relate H to DBH and allow the extrapolation of tree height measurements to all trees in the plots. Then the wood density based on the tree taxonomy was used to calculate the reference aboveground biomass ($\mathrm{AGB}_{ref}$). Aerial LiDAR at a resolution of 1~m was obtained over the sites within an absolute temporal difference of 2.2$\pm$1.9 years from the forest inventory dates. From the LiDAR data canopy height models (CHM) as well as canopy metrics (LCM) were derived. Allometric equations of the form
\begin{equation}
    \label{equ:allometric_equation}
    \log(\mathrm{AGB}_{ref})=a+b\cdot\log(\mathrm{LCM})+\mathrm{RE}_{site}+\epsilon
\end{equation}
were then fit which relate LCM to AGB. The authors found that the average of RH40, RH50, RH60, RH70, RH80, RH90 and RH98 represents the best LCM predictor for AGB.
\\ \\
For this study we utilize the 40~m resolution dataset since it is closer to the native resolution of our model. For the comparison, we resample our models output to 40~m resolution using the average resampling method. We deploy our model over all sites for the year the forest inventory ended at the respective site. Due to limitations of input data availability, our model can only be deployed back to the year 2017. For sites with forest inventory dates prior to 2017, we choose 2017 to be the deployment year. Figure \ref{fig:isro_samples} shows an RGB image of the input data, the reference AGBD as well as our models estimate for AGBD for the site Somaloma, Central Africa.
We consider each pixel of 40~m$\times$40~m size (0.16~ha) as a data point and compare the estimate of AGBD by our model to the ground measurement provided by Rodda et. al. We determine RMSE, ME and MAE across all pixels and site except Achanakmar, Betul and Yellapur which we exclude from this study. For these three sites, we noticed that our model under-estimates AGBD. Upon further investigation, we found the under-estimation to be caused by the input compositing strategy. The three site contain a large fraction of deciduous broadleaf trees and are located in the tropical region for which we construct cloud free composites over a 6 months time period. In these particular cases, this resulted in the inclusion of scenes where the trees did not carry leafs, causing a shift in input signal. This can be addressed by shortening the composite time window for regions with these conditions which we reserve for future work. Figure \ref{fig:isro_results} summarizes the results of this study. The figure on the left shows a 2D histogram of our models estimate vs. the ground measurement for each pixel. There is a clear correlation and we measure an R$^{2}$ value of 0.39. The figure in the middle (right) shows the median error (median absolute error) for each bin of size 5~Mg/ha as well as the interquartile (dark shaded area) and 90~\% (light shaded area) range. These plots show the same pattern as the validation results against the GEDI test dataset (figures \ref{fig:me_binned} and \ref{fig:mae_binned}) where the error is generally small in the range of 0-300~Mg/ha and increases in the region >300~Mg/ha.
This illustrates that there may be limitations related to both the input signal, due to saturation effects, as well as the training data which exhibits increasing uncertainties with larger values of AGBD. The authors of this dataset specifically mention the importance of such ground measurements for re-calibration purposes since the accuracy of AGBD estimates from large scale LiDAR surveys (such as GEDI) is often limited due to the availability of calibration data. In this study, we report the results on testing our model estimates without re-calibration which we plan to do in future work. Overall we measure an ME of -6.46~Mg/ha, a MAE of 60.21~Mg/ha and a RMSE of 84.95~Mg/ha. We also determined the total amount of biomass in all sites considered for this study to be 8.62~Mt based on our models estimate while the total amount based on Rodda et. al. is 8.95~Mt corresponding to a 3.63~\% difference.

\subsubsection{Canopy height}
\label{sec:third_party_canopy_height}
\begin{figure}[h]
	\centering
	\includegraphics[width=0.8\textwidth]{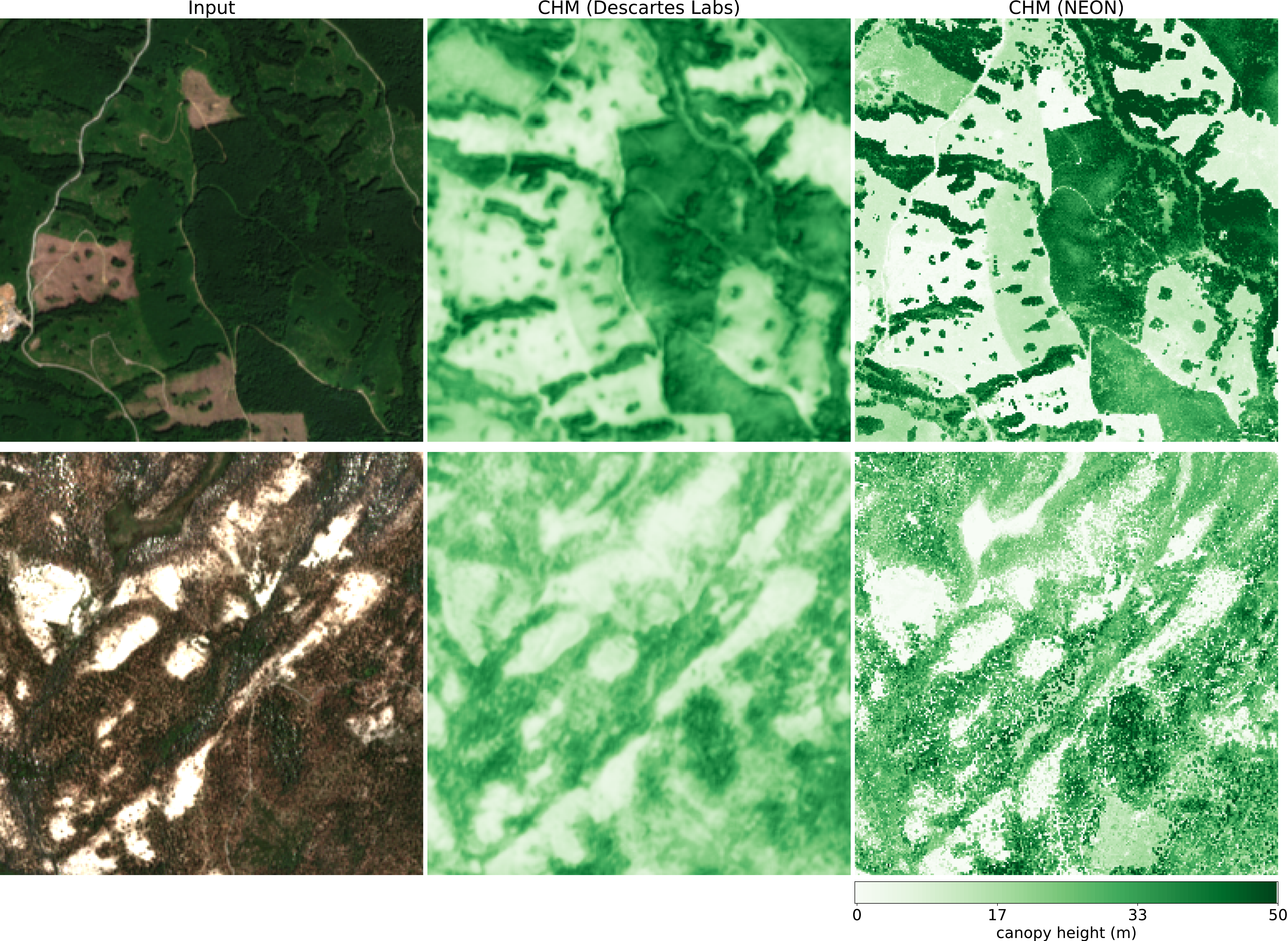}
	\caption{RGB imagery (left), CH estimation of our model (middle) and CH measurements by NEON (right) for two samples within sites ABBY, WA and TEAK, CA.}
	\label{fig:neon_sample}
\end{figure}

\begin{figure}[h]
	\centering
	\includegraphics[width=\textwidth]{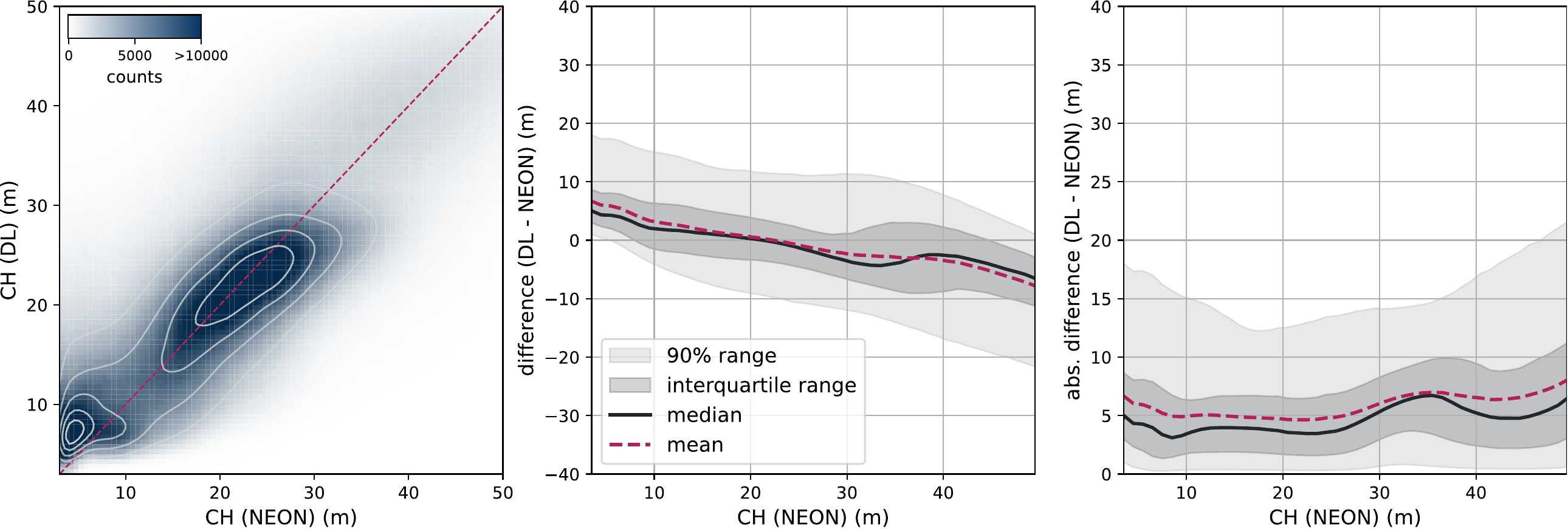}
	\caption{Histogram of estimated CH by our model vs. the measurements by NEON (left), median (middle) and absolute median (right) error as well as means, interquartile and 90~\% ranges in bins of size 1~m with respect to NEON CH.}
	\label{fig:neon_result}
\end{figure}

To further assess our models CH estimation, we use data from the National Ecological Observatory Network (NEON) (\cite{NEON2021}). NEON is a high-resolution LiDAR dataset which provides detailed three-dimensional information about the Earth’s surface. This dataset includes measurements of vegetation structure, topography, and land cover across diverse ecological regions in the United States. The data is collected using airborne LiDAR sensors, capturing fine-scale details at a resolution of 1~m. We selected all measurement sites in the states AL, CA, FL, GA, OR, UT, VA and WA from the year 2021. We subdivide each site into areas of size 2560~m$\times$2560~m and rasterize our models CH estimation as well as NEONs CH measurement at their respective resolution (10~m for our model and 1~m for NEON). This results in tiles of size 256~pixels$\times$256~pixels (our model) and 2560~pixels$\times$2560~pixels (NEON). In order to compare the two maps, we resample the NEON map by determining the 98th percentile in each 10~pixels$\times$10~pixel area resulting in a map of the same pixel size as our models. We choose this approach because our model effectively estimates the RH98 metric for each pixel corresponding to 10~pixels$\times$10~pixels in the NEON map. Figure \ref{fig:neon_sample} illustrates the RGB imagery as well as the CH maps of our model and NEON for two samples (top row corresponds to a sample in site ABBY, WA and bottom row to a sample in site TEAK, CA).
The visual comparison between our models CH estimation and NEONs CH measurement shows very good agreement despite the fact that our models CH map is at 10x lower resolution. We quantify the agreement between our and NEONs CH map by calculating RMSE, ME and MAE considering all pixel values from all tiles and sites as data points. Overall we achieve an RMSE of 7.46~m, an ME of 0.17~m and a MAE of 5.57~m. Figure \ref{fig:neon_result} shows a 2D histogram of all data points as well as ME and MAE aggregated in bins of size 1~m with respect to the NEON CH. The R$^{2}$ value of the model predictions compared to NEONs measurements is 0.51. We also illustrate the error distribution within each bin by indicating the interquartile as well as 90~\% range. The red, dashed line corresponds to the mean in each bin.
The agreement between our and NEONs CH map is very good across the entire value range. We note that there is some larger disagreement at low values (<5~m) which is also visible in the sample maps in figure \ref{fig:neon_sample} where our model generally estimates slightly higher values at small CH. This can be attributed to the fact that the GEDI dataset contains a very small amount of data points <3~m so the model generally over-estimates in this region. However, it is remarkable that the absolute error mostly stays below 5~m across the entire value range up to 50~m. The NEON dataset offers the capability to re-calibrate the model estimations where it exhibits a bias. This is true for any high-accuracy, regional dataset. We reserve the demonstration of re-calibration of the model for future work.

\section{Applications}
\label{sec:applications}
In this section we demonstrate some use cases of our model including the monitoring of changes over time as well as fine-tuning the model on regional ground truth data to better align the global model to local conditions. Traditionally, many use cases require the deployment of multiple models in order to perform a down-stream task where each model output provides the solution for sub-tasks. One of the main motivations of this work is to create a single model which is versatile enough to provide results for all sub-tasks in one forward pass. One such application is the detection of deforestation and association of loss of biomass which we discuss in detail in section \ref{sec:monitoring_change}.

\subsection{Global model deployment}
\label{sec:global_model_deployment}
\begin{figure}[h]
	\centering
            \begin{subfigure}{0.8\textwidth}
                \centering
	    \includegraphics[width=\textwidth]{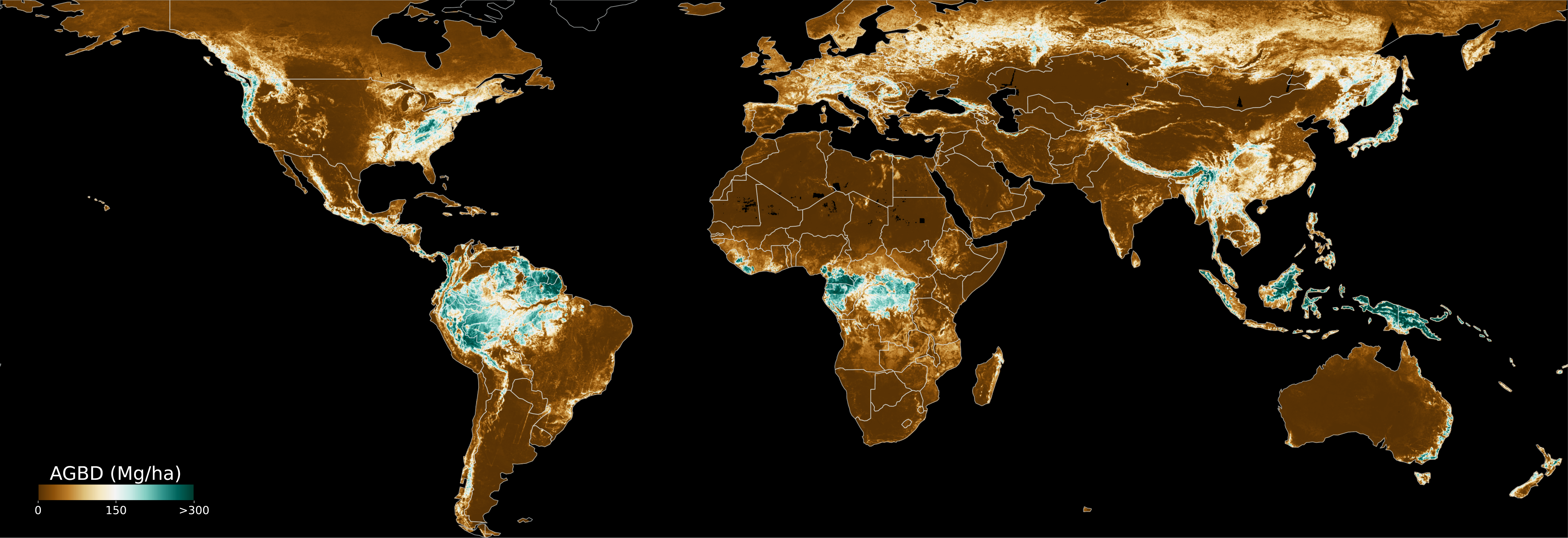}
	       \caption{Aboveground biomass density}
            \end{subfigure}
            \begin{subfigure}{0.8\textwidth}
                \centering
            \includegraphics[width=\textwidth]{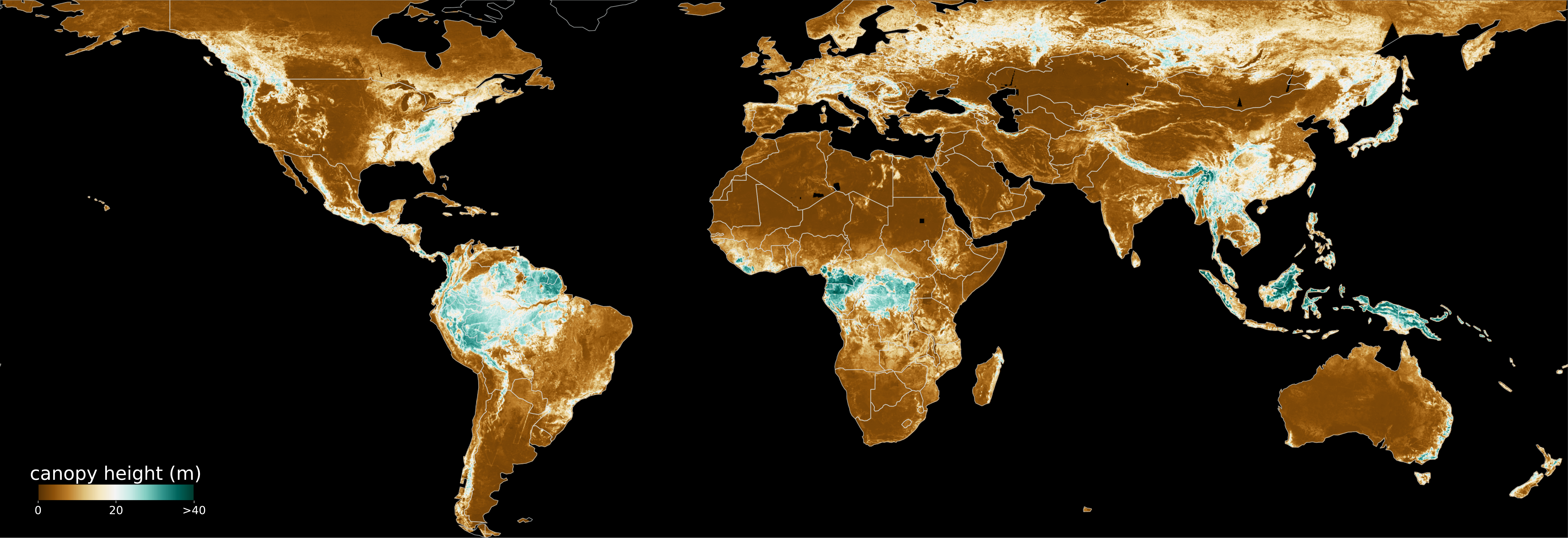}
	       \caption{Canopy height}
            \end{subfigure}
            \begin{subfigure}{0.8\textwidth}
                \centering
            \includegraphics[width=\textwidth]{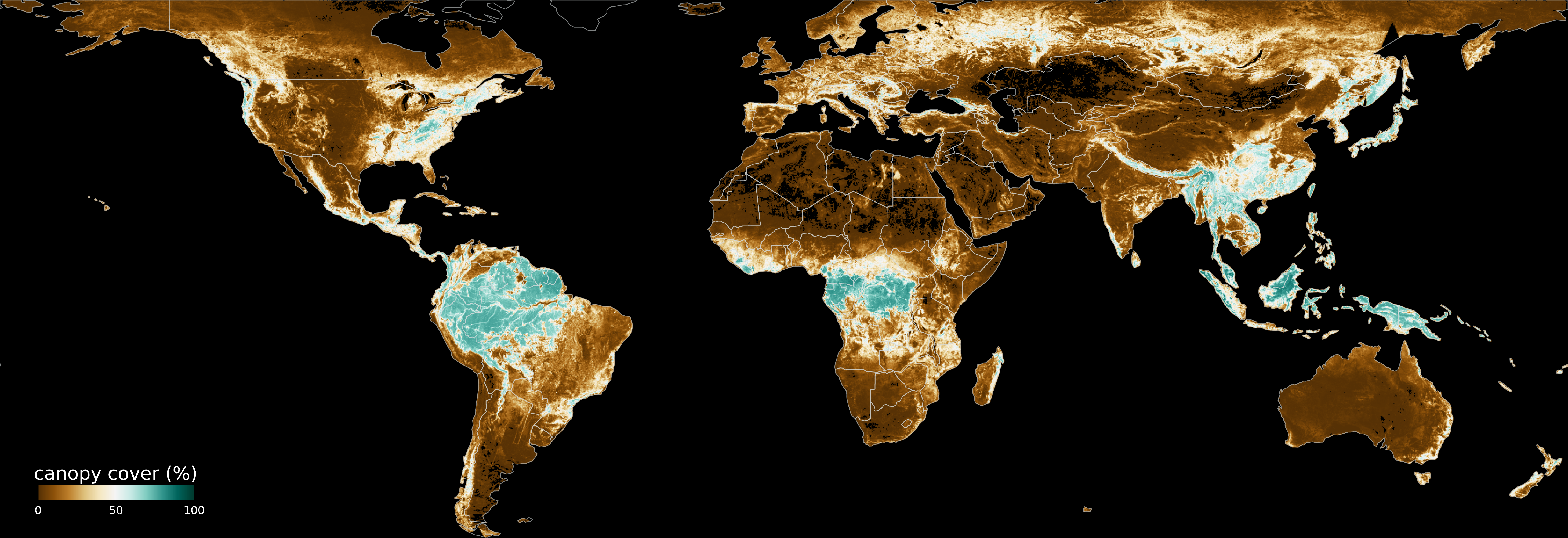}
	       \caption{Canopy cover}
            \end{subfigure}
        \caption{Global maps of aboveground biomass density, canopy height and canopy cover at 10~m resolution for the year 2023}
	\label{fig:world_maps}
\end{figure}

Our model is deployable at scale due to its global training dataset. This is an important factor when it comes to monitoring changes in the ecosystems. Our model is scalable both spatially as well as temporally which is important for change detection. While global deployments may not be required very frequently, deployments over local regions of interest can rapidly be performed at temporal resolutions of <1 year.
\\ \\
We generated global maps for all three prediction variables AGBD, CH and CC as well as their uncertainties at 10~m resolution for the recent years. They cover the latitude range of 57$^{\circ}$~S to 67$^{\circ}$~N according to the availability of Sentinel-1 data. The deployment year is 2023 for all regions except for the areas where Sentinel-1 experienced an outage for that year. In this case, we filled the gaps with the most recent available observations which, in most cases, is 2021. The deployment was performed in two steps utilizing AWS scalable Batch compute: First we generate cloud free composites, pulling data from the Descartes Labs platform, for global image tiles of size 1024~pixels$\times$1024~pixels. The tiles were generated using the Descartes Labs tiling system. We follow the same schema for determining the composite time window as for the training dataset (see section \ref{sec:global_dataset}). The image composites are saved as product in the Descartes Labs platform for fast retrieval during inference. In the second step, we run model inference on tiles of size 2048~pixels$\times$2048~pixels with a padding of 80 pixels. The output of the deployment is stored as product in the Descartes Labs platform and consists of multiple layers representing the prediction variables as well as multiple mask layers. The mask layers are either forward propagated from the composite product, such as composite gap mask, or derived from the model output, such as forest mask. Figure \ref{fig:world_maps} shows the low resolution maps for all prediction variables.

\subsection{Monitoring biomass change and deforestation detection}
\label{sec:monitoring_change}
\begin{figure}[h]
	\centering
	\includegraphics[width=0.8\textwidth]{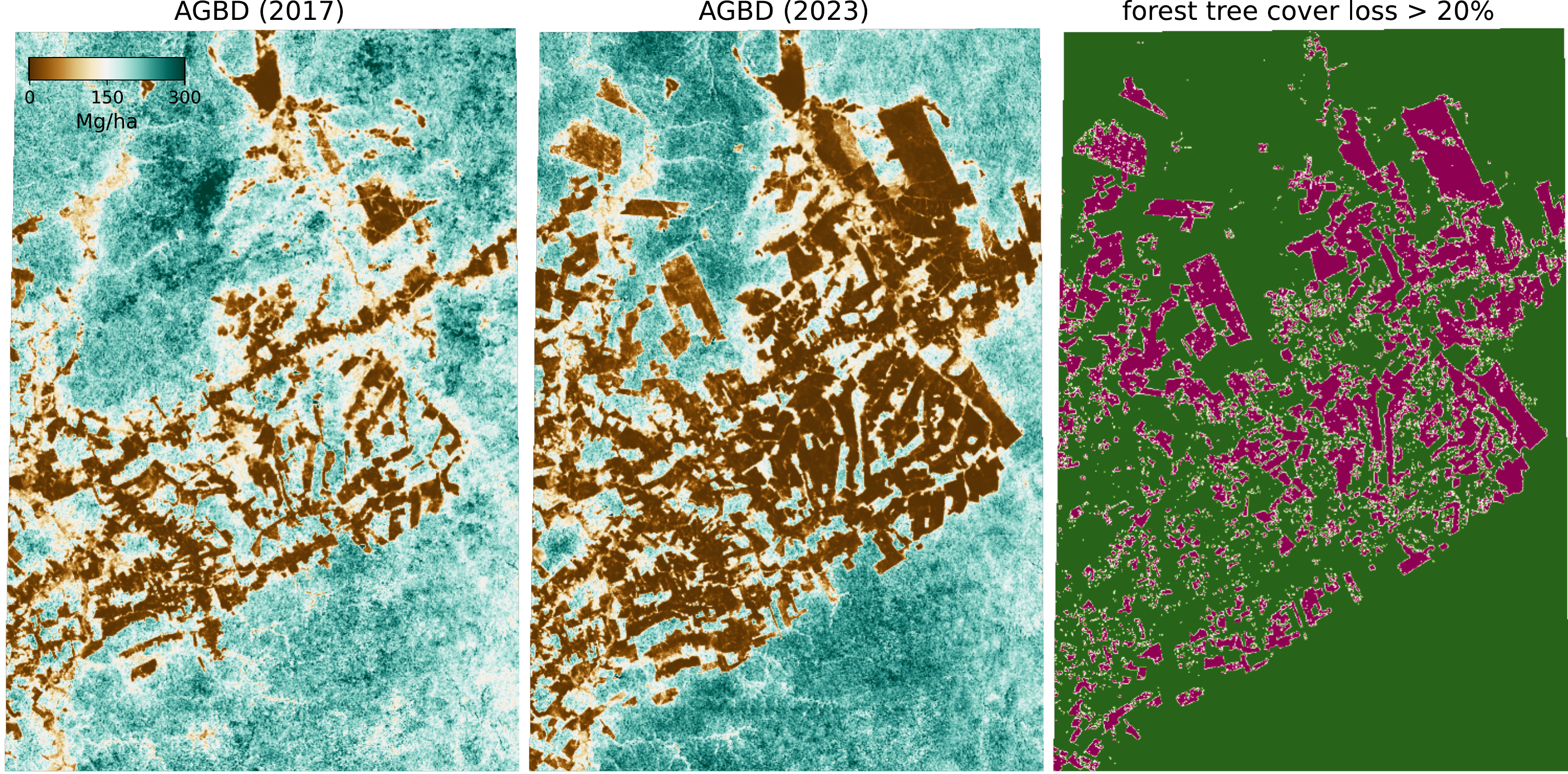}
	\caption{Map of above ground biomass density in 2017 (left) and 2023 (right) for a selected area of interest in Brazil as well as areas of significant change in canopy cover (right).}
	\label{fig:deforestation_map}
\end{figure}

\begin{figure}[h]
	\centering
	\includegraphics[width=0.6\textwidth]{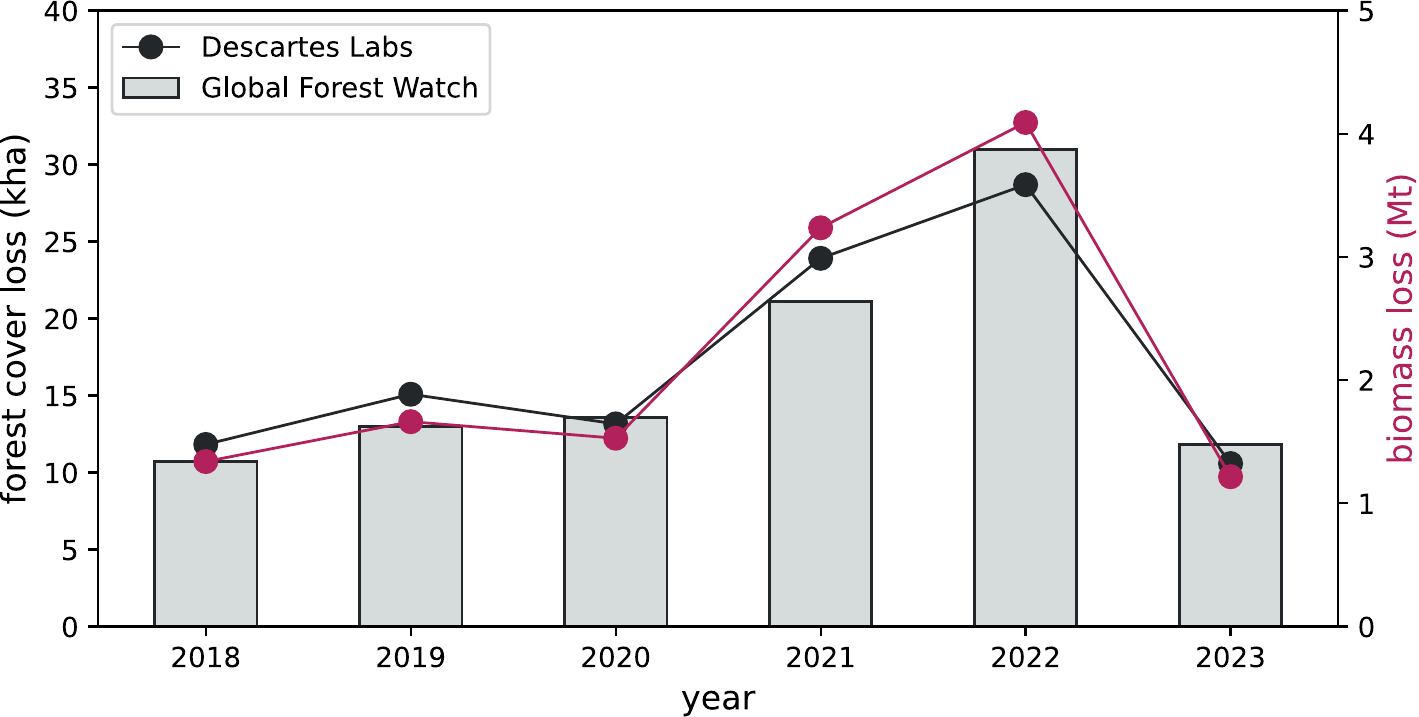}
	\caption{Annual tree cover loss area for a selected AOI in Brazil and comparison to reports from Global Forest Watch.}
	\label{fig:deforestation}
\end{figure}

The monitoring of global deforestation and carbon accounting has become an important part of climate change mitigation where remotely sensed data plays an integral part for many models developed in the past. Most these models utilize optical or SAR based imagery and are based on a variety of approaches, from pixel based anomaly detection in time series data (\cite{DECUYPER2022112829}) to modern deep learning and computer vision algorithms (\cite{SOLORZANO202387}). In this section we demonstrate the usability of our model for the task of deforestation detection and accounting of corresponding loss in biomass. We selected an area in Brazil at location 59$^{\circ}$~26'~46''~W, 7$^{\circ}$~1'~4''~S with a total area of 507~kha. We deploy our model over the area for each year from 2017 to 2023. Figure \ref{fig:deforestation_map} shows the map of AGBD for 2017 (left) and 2023 (middle) respectively. In this study, we measure changes year over year and compare the results to Global Forest Watch (\cite{HANSEN2013}) which provides annual tree cover loss data based on observations from Landsat at 30~m resolution. We follow their approach for defining forested land by requiring CH > 5~m and determine the regions with significant canopy cover loss ($\Delta$CC > 20~\%). Figure \ref{fig:deforestation_map} (right) shows the regions which meet these requirements accumulated from 2017 to 2023. Note that not all regions correspond to complete deforestation but rather significant tree cover loss in accordance to the definition by Global Forest Watch. We aggregate the areas with tree cover loss for each year, with respect to the previous year, and compare our numbers to those reported by Global Forest Watch. Figure \ref{fig:deforestation} summarizes the results.
Our numbers agree with those by Global Forest Watch to within 10~\% on average. We determine the total area of tree cover loss over the 6 year time span to be 103.3~kha while Global Forest Watch reports a number of 101.2~kha, a difference of 2.1~\%. In addition to the total area of tree cover loss, our model also provides the amount of biomass lost in these areas by taking the difference of AGBD maps in 2017 and 2023 and multiplying by the total area. We measure a total loss in biomass of 14.9~Mt, equivalent to 25.66~Mt of CO$_{2}$.
This study highlights the practicality of our model as a unified approach for simultaneous prediction of multiple relevant variables and it makes it easily scalable to other regions in the world.

\subsection{Model fine-tuning for local conditions}
\label{sec:model_finetuning}
\begin{figure}[h]
	\centering
	\includegraphics[width=\textwidth]{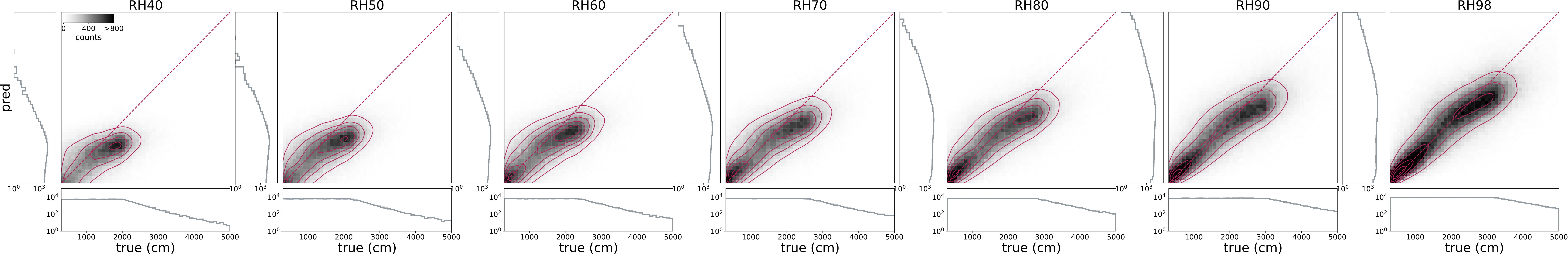}
	\caption{Histograms of predicted vs. true values for all RH metrics that the model was fine-tuned on.}
	\label{fig:y_pred_y_true_LCM}
\end{figure}

\begin{figure}[h]
	\centering
	\includegraphics[width=0.8\textwidth]{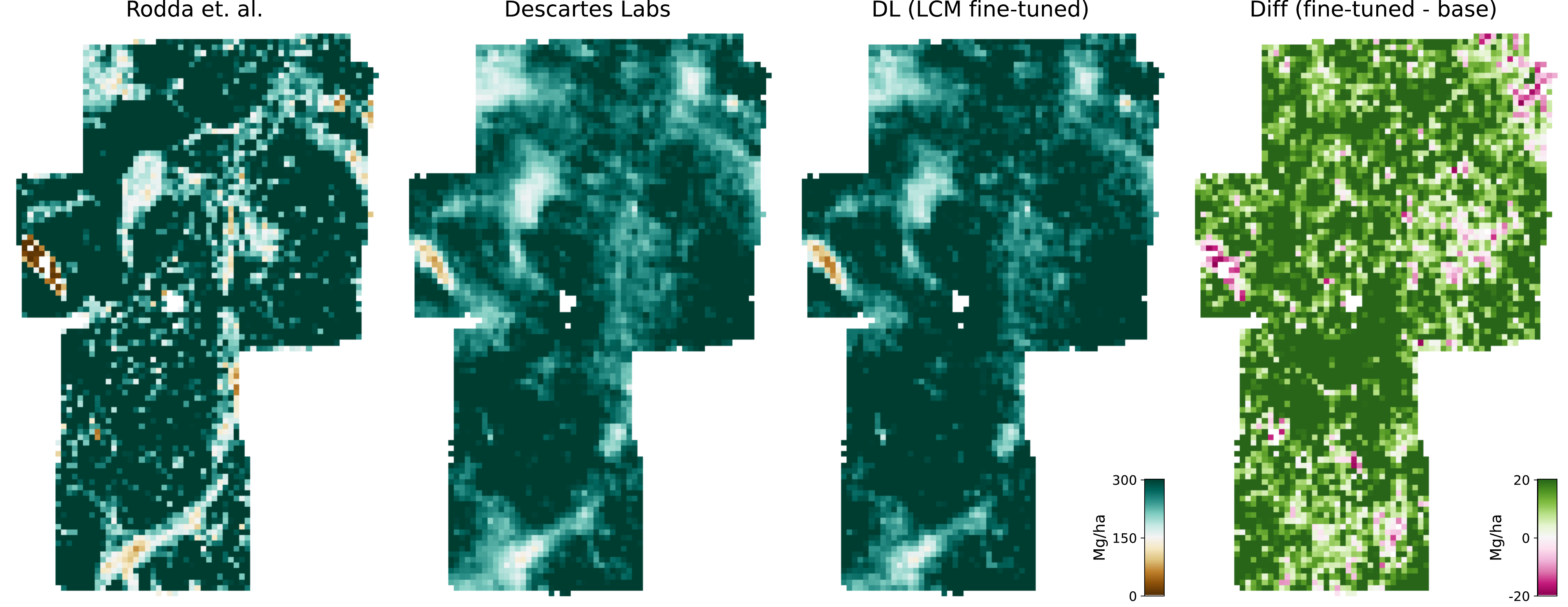}
	\caption{Comparison of AGBD maps from Rodda et. al. (left), our base model (middle) and our fine-tuned model for RH variables and local allometric equations (right).}
	\label{fig:isro_samples_lcm}
\end{figure}

One of the main contributors to the uncertainty of AGBD estimations are limited availability of ground measurements for calibrating allometric equations at global scale. On the other hand, estimations of canopy height metrics are not depending on such calibrations as they are derived from the LiDAR waveforms directly. Our model therefore offers a unique capability for fine-tuning on local conditions in order to achieve more accurate AGBD estimations. The majority of allomtric equations use relative height (RH) metrics as predictors for AGBD which our model is able to estimate. The base model predicts AGBD, RH98 and CC. However, due to its multi-head architecture, it can easily be fine-tuned on other variables. We demonstrate such a use case based on the data and coefficients for allometric equations provided by (\cite{RODDA2024}). The authors determine the coefficients of the allometric equation in \ref{equ:allometric_equation} for each of the 13 sites. They find that the average of all listed RH metrics is the best predictor for AGBD. We expand our model to 7 heads, each representing one of the RH metrics and fine-tune the weights of these heads for one epoch while keeping the weights of the encoder and decoder frozen. We apply sample weighting according to the inverse PDF of each variables distribution (see appendix \ref{apx:uniform_sampling} for details). Figure \ref{fig:y_pred_y_true_LCM} shows 2D histograms of the models estimate for each variable versus the true value, evaluated on a uniformly sampled test dataset.
Generally there is very good agreement between the models predictions and the true values for all RH metrics. The detailed evaluation metrics are summarized in table \ref{tab:result_lcm}. We deployed the fine-tuned model over the sites given by Rodda et. al. using the same input data as in section \ref{sec:third_party_biomass} and determine AGBD based on the site specific allometric equations and the RH predictions of our model. Figure \ref{fig:isro_samples_lcm} shows AGBD maps for the site Somalomo, Central Africa, based on Rodda et. al. (left), AGBD estimate from our base model (mid-left), AGBD derivation from the RH variables of our fine-tuned model (mid-right) and the difference between the fine-tuned and the base model (right). Visually, there are subtle differences between our base- and fine-tuned model. However, the fine-tuned model decreases the MAE by 15.3~\% to 50.96~Mg/ha and the RMSE by 16.2~\% to 71.18~Mg/ha. The R$^{2}$ value increases from 0.39 to 0.58.
\begin{table}[h!]
\centering
\begin{tabular}{c|c|c|c|c|c|c|c}
\hline \hline
metric & RH40 & RH50 & RH60 & RH70 & RH80 & RH90 & RH98 \\
\hline
ME & -327.48 & -283.10 & -235.92 & -185.50 & -152.69 & -112.22 & -91.73 \\
MAE & 511.92 & 506.91 & 497.92 & 490.29 & 484.57 & 478.49 & 473.47 \\
RMSE & 653.74 & 657.18 & 654.83 & 652.30 & 648.12 & 647.48 & 642.94 \\
corr & 0.559 & 0.608 & 0.657 & 0.698 & 0.736 & 0.768 & 0.800 \\
\hline \hline
\end{tabular}
\caption{Summary of evaluation metrics for all RH variables from the fine-tuned model. All metrics are given in unit cm.}
\label{tab:result_lcm}
\end{table}
For more details on the comparison between the base- and fine-tuned model, see appendix \ref{apx:assessment_lcm}. These results highlight the benefit of local model fine-tuning in order to incorporate regional calibrations of allometric equations which our model is inherently designed for.

\section{Discussion}
\label{sec:discussion}
In this work we propose a novel deep learning based model which unifies the prediction of several biophysical indicators which describe the structure and function of vegetation across multiple ecosystems. We utilize both Sentinel-1 backscatter as well as multi-spectral Sentinel-2 data at 10~m resolution. We further enrich the input data by adding topographic information from the SRTM dataset as well as geographic coordinates. We developed an approach which allows the end-to-end training of the model through a weakly supervised learning method on point data representing individual measurements of the respective variables. For this purpose, we construct a loss function which balances the varying number of point data per sample as well as the distribution non-uniformity and simultaneously allows the estimation of the variable uncertainty. We further demonstrate how the resulting base model can easily be fine-tuned on any number of variables contained in the ground truth dataset. We perform a rigorous evaluation procedure to assess the models capability against a held out test dataset sampled at global scale. We find that the model performs very well against the test dataset with an RMSE of 50.59~Mg/ha (543.81~cm, 15.75~\%) for AGBD (CH, CC), outperforming similar state-of-the-art models such as (\cite{Bereczky2024}) for AGBD and (\cite{pauls2024estimatingcanopyheightscale}) for CH. We further evaluated our model against third party datasets without further fine-tuning and attained performance that is consistent with or surpasses the expectations for such data, demonstrating the robustness and generalizability of our approach. As with previous works, we notice a saturation effect at higher values of AGBD and CH which can be attributed to both a lack of training data in those regions and a saturation at the input level. This is an inherent limitation when using remotely sensed data as the information content is restricted. The effect is more severe with AGBD which has an additional component from the increasing uncertainty of ground truth samples at higher values due to the errors propagated from the calibration.
\\ \\
We demonstrate the scalability of our model, due to its global training dataset, by generating global maps of AGBD, CH and CC at 10~m resolution, extending the coverage of GEDI observations to a latitude range of 57$^{\circ}$~S to 67$^{\circ}$~N. We show how the fusion of multiple input sources is beneficial in areas of imperfect cloud free composites, a limitation which previous works have experienced. These maps represent the most up-to-date data available at the time of publishing by using input data from the year 2023 or 2021 respectively where Sentinel-1 data was unavailable.

\section{Acknowledgement}
\label{sec:acknowledgement}
We would like to thank Piyush Agram, Scott Arko, Rachel Landman and Jacob McKee from the Descartes Labs data ingest team who have contributed to this work by creating data processing pipelines to make the various data components available through the Descartes Labs platform. We would also like to acknowledge the hard work of the entire engineering team at Descartes Labs who built and maintain the platform and made it possible to leverage its scalable compute for both generating training data and global deployment.

\appendix

\section{Uniform data sampling}
\label{apx:uniform_sampling}
\begin{figure}[h]
	\centering
            \begin{subfigure}{0.3\textwidth}
                \centering
	    \includegraphics[width=\textwidth]{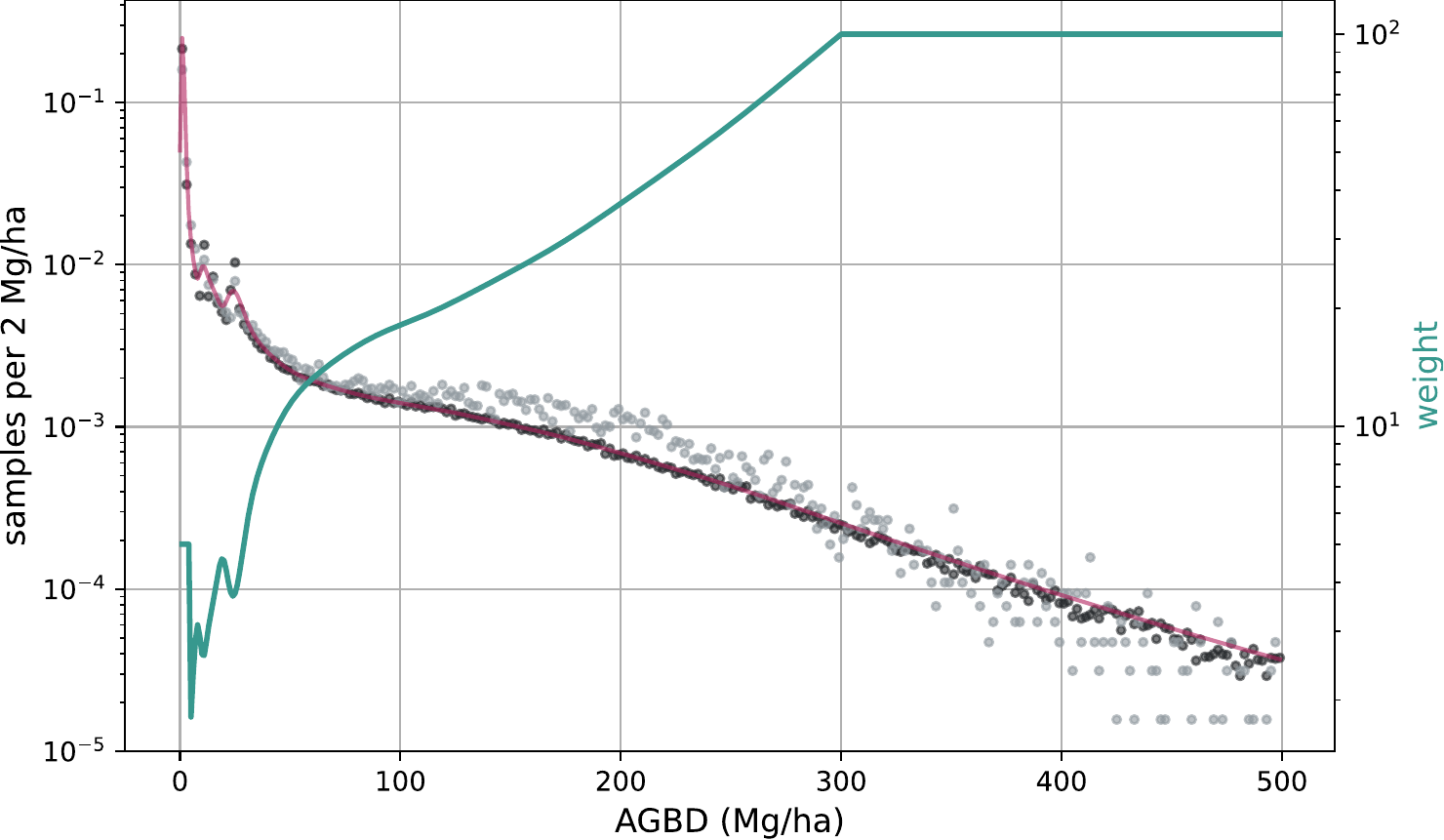}
	       \caption{Aboveground biomass density}
            \end{subfigure}
            \begin{subfigure}{0.3\textwidth}
                \centering
            \includegraphics[width=\textwidth]{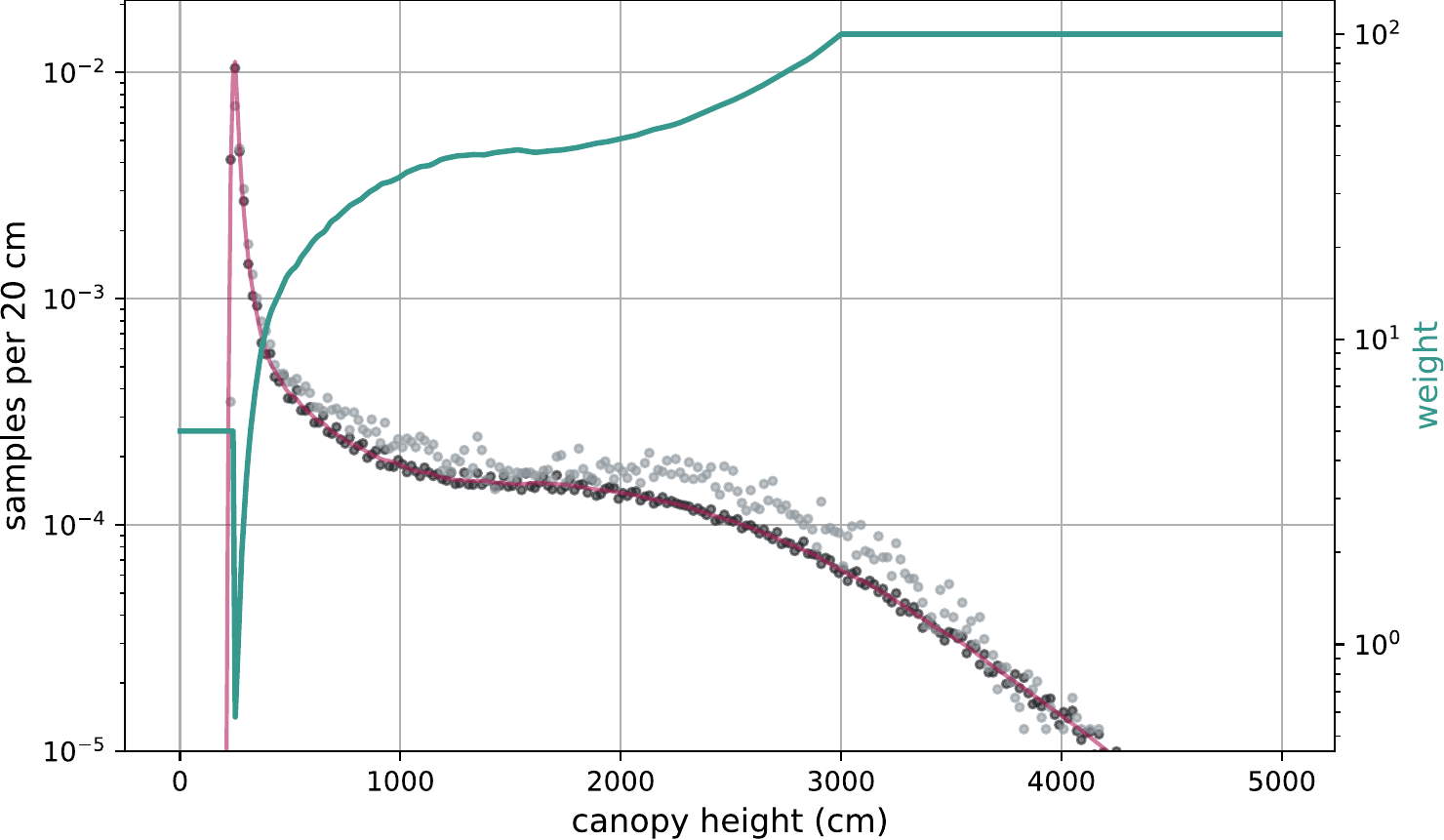}
	       \caption{Canopy height}
            \end{subfigure}
            \begin{subfigure}{0.3\textwidth}
                \centering
            \includegraphics[width=\textwidth]{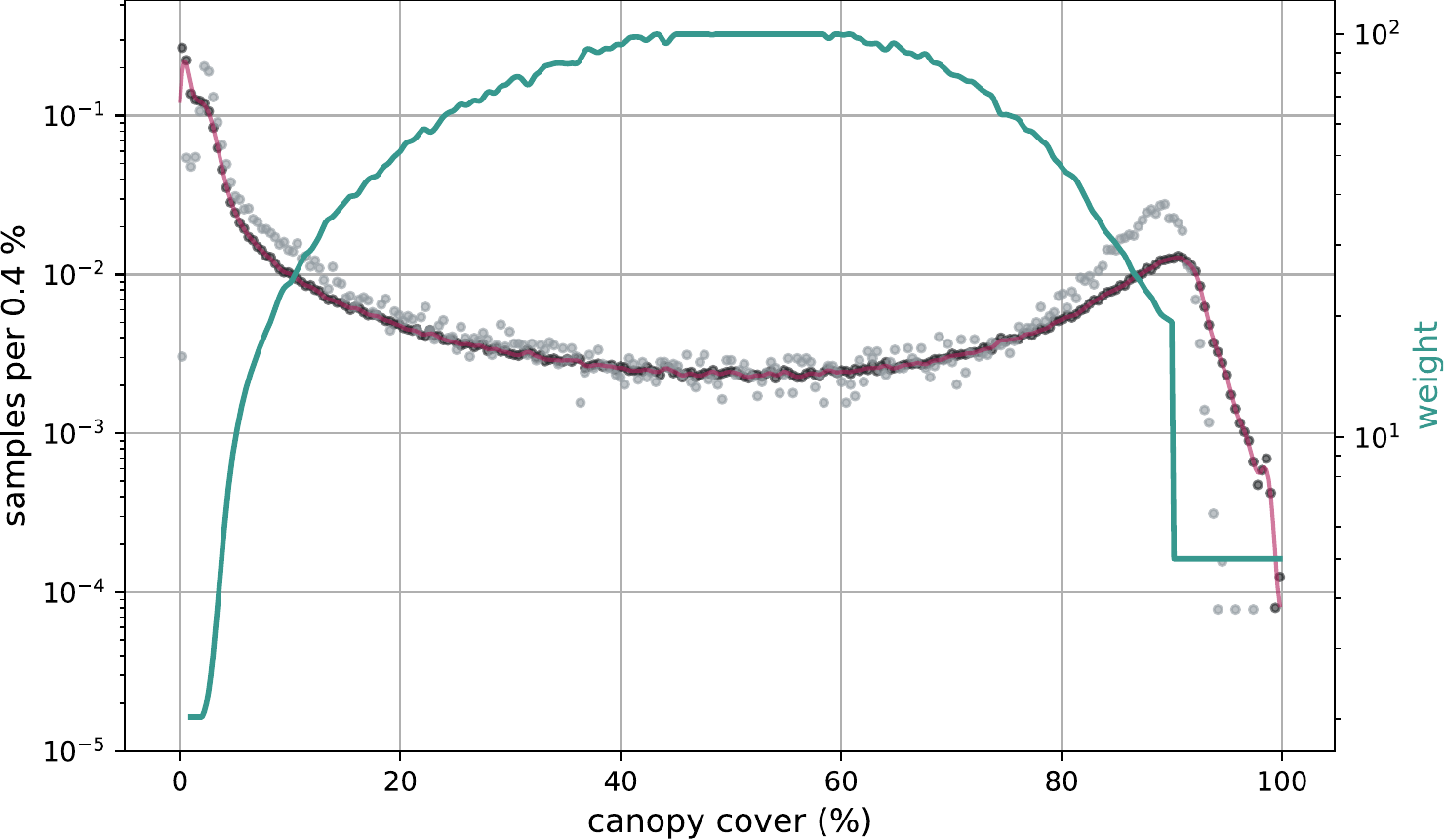}
	       \caption{Canopy cover}
            \end{subfigure}
        \caption{Binned sample distributions for each variable as well as corresponding distribution of 75th percentile of all values within a given tile, the fitted kernel density estimation as well as weight function.}
	\label{fig:sample_distributions}
\end{figure}

\begin{figure}[h]
	\centering
	\includegraphics[width=\textwidth]{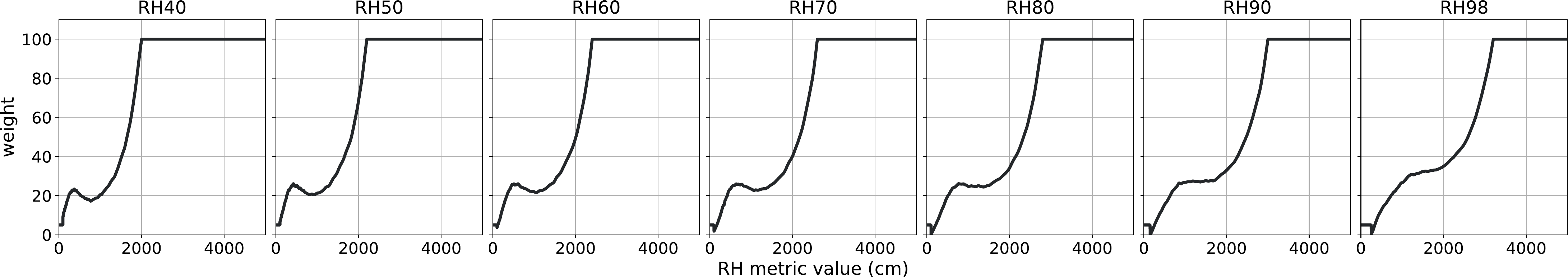}
	\caption{Weight functions of all RH variables used in the fine-tuning of the model.}
	\label{fig:sample_distributions_lcm}
\end{figure}

All target variables used to train the model have a highly non-uniform distribution of sample values. Without sample weighting, this can lead to over- (under-)fitting in regions with higher (lower) sample frequency. In order to account for this imbalance, we determine variable specific weight functions according to the inverse probability distribution function (PDF) of the sample distributions. The PDFs are determined by kernel density estimations (KDE) of the binned data distributions. Figure \ref{fig:sample_distributions} shows the sample distributions of ABGD (a), CH (b) and CC (c). The fitted KDE is shown as purple line while the resulting weight function is shown as green line. The grey data points correspond to the distribution of the 75th percentile of values within a given tile. This distribution is used for creating a balanced dataset (see section \ref{sec:training_details}). The weight function is incorporated into the loss function \ref{eq:balanced_loss} as described in section \ref{sec:loss_function}. During training, the weight function needs to be evaluated for each sample in the batch which can be computationally intensive when using a KDE. We therefore save the weight functions as pre-computed lookup tables which can be evaluated with minimal time delay.
The same procedure is applied for fine-tuning the model on RH variables (see section \ref{sec:model_finetuning}) where the weight functions are determined for each variable separately. Figure \ref{fig:sample_distributions_lcm} illustrates the weight functions for all variables.

\section{Model performance details}
\label{apx:model_performance_details}
This section contains further details on the various studies conducted for assessing the model performance.

\subsection{Coverage of uncertainty estimations}
\label{apx:uncertainty_estimation}
\begin{figure}[h]
	\centering
	\includegraphics[width=0.8\textwidth]{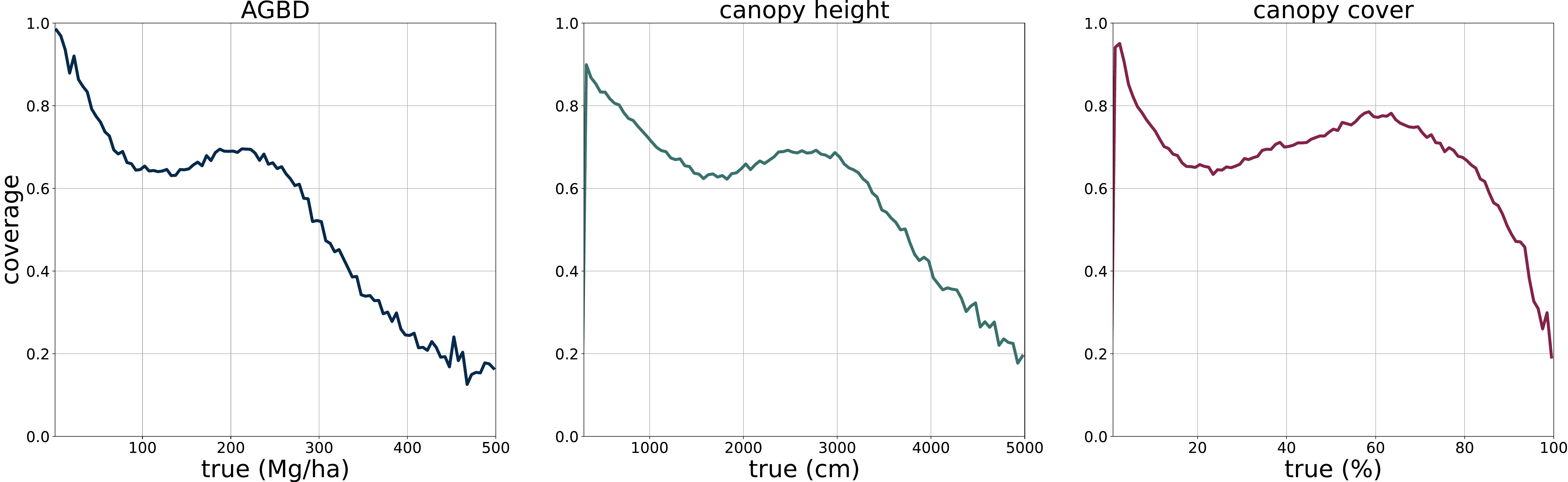}
	\caption{Coverage, defined as the fraction of samples with a z-score <1, as a function of true value for all variables.}
	\label{fig:coverage}
\end{figure}

Our model estimates the uncertainty of each variable with additional prediction heads and incorporating the variances in the loss function \ref{eq:loss_regularized}. Whether the model correctly predicts the standard error can be verified by measuring the fraction of samples with a z-score <1 which is expected to be 68~\%. During training, we choose the regularization weight in equation \ref{eq:loss_regularized} such that this fraction reaches 0.68 across the full variable range. However, the fraction may vary across different values of the predicted variable. Figure \ref{fig:coverage} illustrates the z-score fraction as a function of the true value in bins of 5~Mg/ha for AGBD, 50~cm for CH and 1~\% for CC. It shows that the standard error is slightly over-estimated for lower values, leading to coverage >68~\%, and tends to be under-estimated at higher values, causing the coverage to be <68~\%. We will address this discrepancy in future works. However, these plots can be used in order to interpret the models uncertainty estimations at a given variable prediction.

\subsection{Effect of target uncertainties on predictions}
\label{apx:gedi_uncertainty}
\begin{figure}[h]
	\centering
	\includegraphics[width=0.8\textwidth]{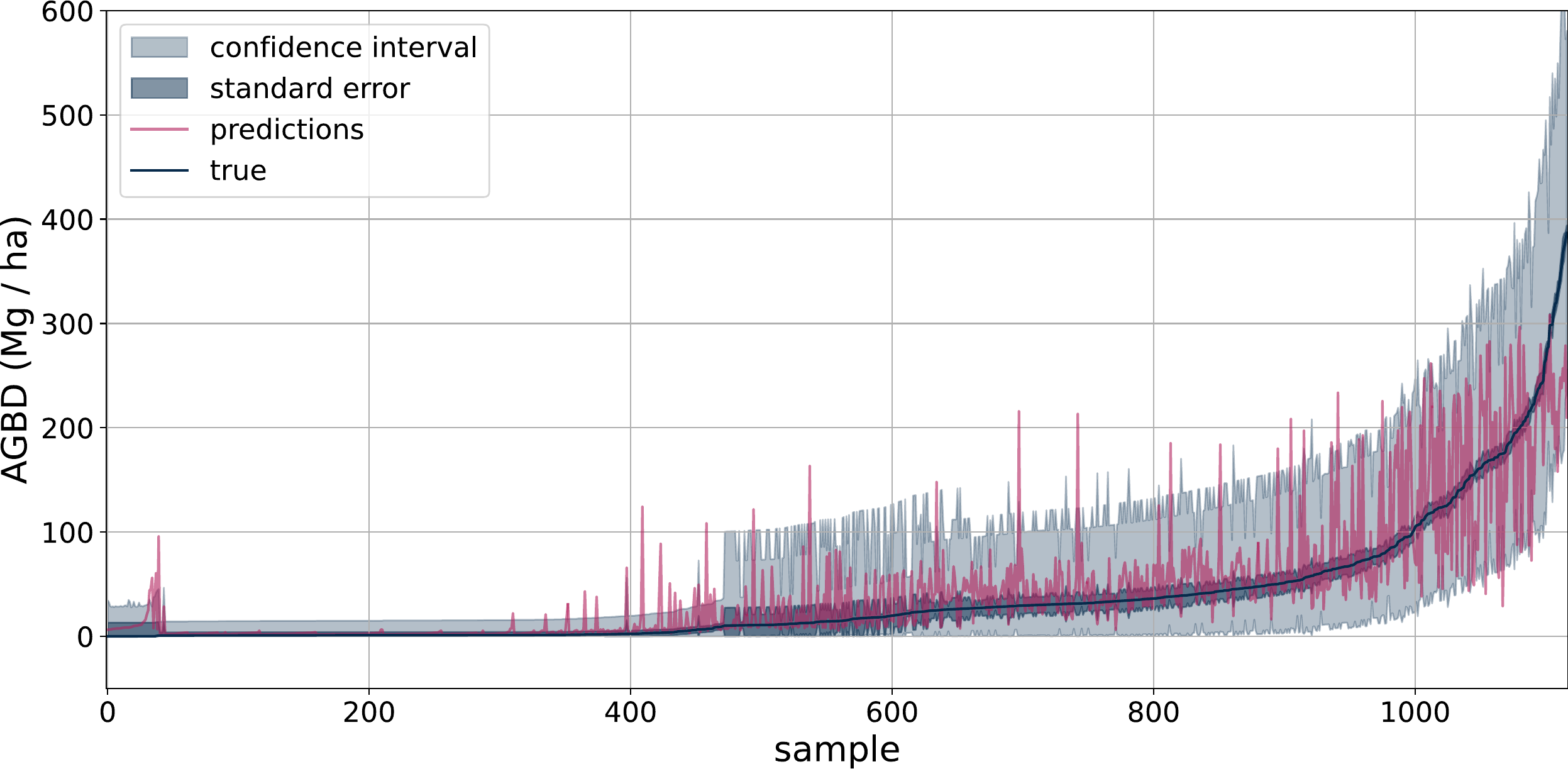}
	\caption{Predicted and true values of AGBD as well as the standard error and 95~\% confidence interval for each sample in a given batch, ordered by the true value from low to high.}
	\label{fig:gedi_uncertainty}
\end{figure}

In general, when fitting a model to observed data, the inherent uncertainty on each data point can be accounted for by proper construction of the objective function such as the reduced $\chi^{2}$ which then allows to determine the uncertainties imposed on each parameter of the model. However, when the model is represented by a deep neural network, the incorporation of ground truth uncertainties is much more complicated as the fitting procedure is done iteratively with mini batches and the number of parameters is far bigger than usual models characterizing a physical process. During training, each target sample is essentially considered to be the true value despite its uncertainty. This can lead to situations where the same input X is associated to two different outputs $\mathrm{Y}_{1}$ and $\mathrm{Y}_{2}$. This can not be modeled by a continuous function which is why the model likely learns the prediction of the average between $\mathrm{Y}_{1}$ and $\mathrm{Y}_{2}$. Therefore, the predictions will also have an uncertainty but it is not trivial to quantify them. The ground truth values for AGBD have an uncertainty attached due to the uncertainties of the predictors in the allometric equations \ref{eq:gedi_agbd_se}. These uncertainty quantifications are provided in the GEDI level-4 dataset as the standard error and 95~\% confidence intervals. In order to investigate the effect of these uncertainties on the models prediction of AGBD, we plot the true and predicted values for all samples within a given batch ordered by the true value from small to large. Figure \ref{fig:gedi_uncertainty} shows the result together with the uncertainty and confidence intervals.
The predictions follow the trend of the true values but fluctuate around the true value. This fluctuation is larger as the uncertainty interval increases which happens at higher values of AGBD. At lower values, the uncertainties are much smaller which causes the predictions to have smaller variance. At very small values, the uncertainty increases again, causing higher prediction variance. This pattern confirms that the uncertainty on the ground truth has a clear effect on the uncertainty of the predictions. It is therefore reasonable to assume that the saturation of the models predictions at higher values of AGBD is partly caused by the uncertainties in the ground truth samples.

\subsection{Performance assessment for different PFTs}
\label{apx:assessment_pft}
\begin{figure}[h]
	\centering
	\includegraphics[width=0.8\textwidth]{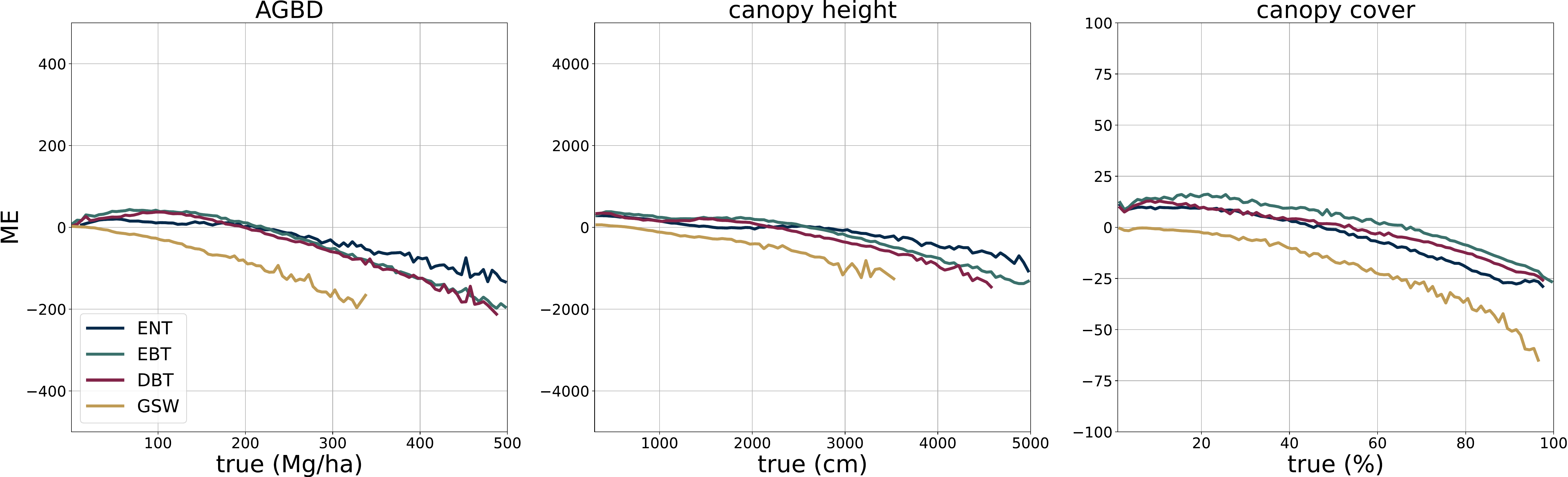}
	\caption{Median of errors between predicted and true values in each bin vs true value for the variables AGBD (left), CH (middle) and CC (right) separated by plant functional type.}
	\label{fig:me_binned_pft}
\end{figure}

\begin{figure}[h]
	\centering
	\includegraphics[width=0.8\textwidth]{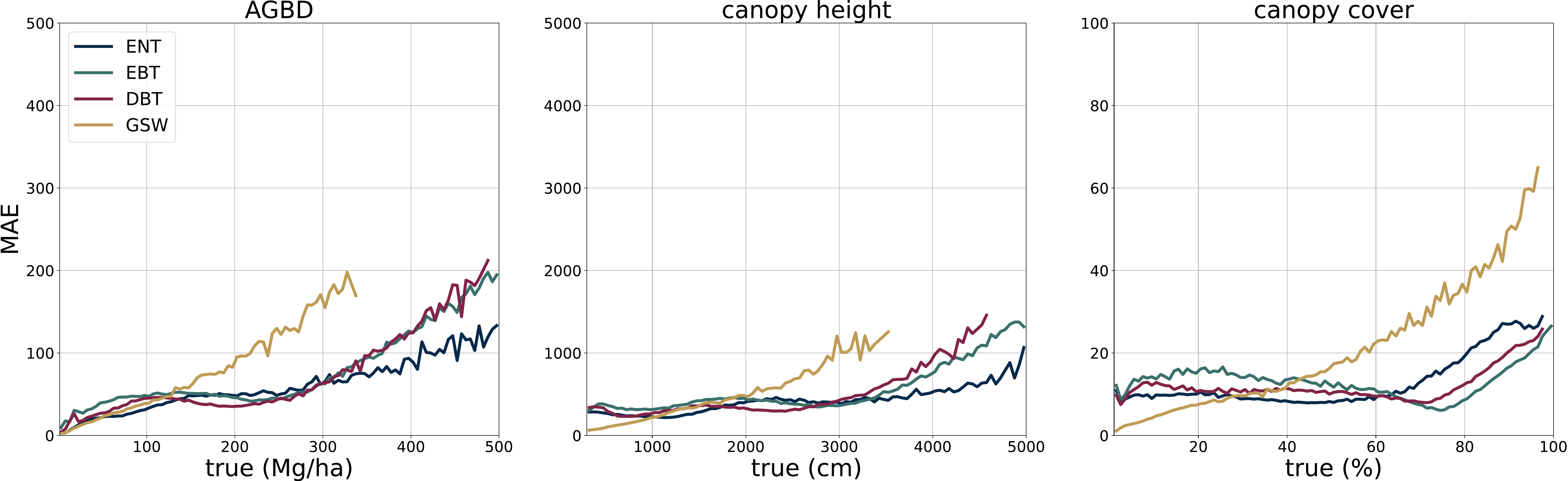}
	\caption{Median of absolute errors between predicted and true values in each bin vs true value for the variables AGBD (left), CH (middle) and CC (right) separated by plant functional type.}
	\label{fig:mae_binned_pft}
\end{figure}

The model assessment in section \ref{sec:quantitative_assessment} was performed across all world regions and plant functional types (PFT). We repeated the analysis where we grouped the GEDI data point according to their PFT classification. This classification is provided within the GEDI dataset. We use the following classes: Deciduous Broadleaf Trees (DBT), Evergreen Broadleaf Trees (EBT), Evergreen Needleleaf Trees (ENT) and Grasslands/Shrublands/Woodlands (GSW).
\begin{table}[h!]
\centering
\begin{tabular}{c|c|c|c|c}
\hline \hline
variable & DBT & EBT & ENT & GSW \\
\hline
AGBD & 66.98 & 84.33 & 66.53 & 24.30 \\
CH & 585.31 & 700.01 & 608.94 & 318.72 \\
CC & 21.82 & 21.53 & 21.19 & 9.05 \\
\hline \hline
\end{tabular}
\caption{RMSE for all variables based on plant functional types (PFT). The units for AGBD (CH, CC) are Mg/ha (cm, \%).}
\label{tab:result_summary_pft}
\end{table}
Table \ref{tab:result_summary_pft} summarizes the determined values of RMSE for all variables and PFTs. Figures \ref{fig:me_binned_pft} and \ref{fig:mae_binned_pft} show the median error and median absolute error for each bin, separated by the different PFTs. For these plots we only include bins with >50 entries/bin.

\subsection{Model fine-tuning for local RH metrics}
\label{apx:assessment_lcm}
\begin{figure}[h]
	\centering
	\includegraphics[width=0.8\textwidth]{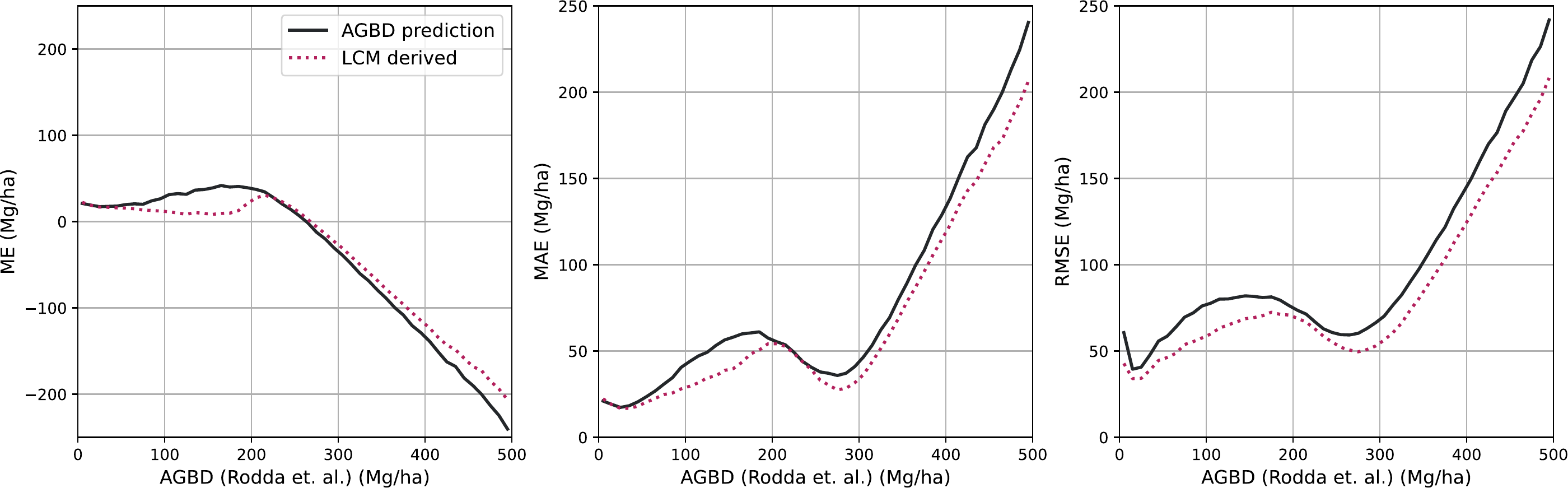}
	\caption{Comparison of evaluation metrics between the base model which predicts AGBD directly and the fine-tuned model which predicts RH metrics and uses locally calibrated allometric equations.}
	\label{fig:isro_results_lcm}
\end{figure}

Figure \ref{fig:isro_results_lcm} shows a comparison of ME, MAE and RMSE metrics between the base- and fine-tuned model. The base model predicts AGBD directly while the fine-tuned model predicts RH metrics which serve as predictors for locally calibrated allometric equations. Using the fine-tuned model decreases the overall error w.r.t the base model. The evaluation metrics are given as a function of the ground measurements by Rodda et. al. in bins of size 10~Mg/ha.

\bibliographystyle{unsrtnat}
\bibliography{references}

\end{document}